%% file: frrn_paper.tex
\documentclass[10pt,twocolumn,letterpaper]{article}

\usepackage[dvipsnames,table]{xcolor}
\usepackage{subcaption}
\usepackage{cvpr}
\usepackage{times}
\usepackage{epsfig}
\usepackage{graphicx}
\usepackage{amsmath}
\usepackage{amssymb}
\usepackage{marvosym}
\usepackage{booktabs}

\usepackage{csvsimple}
\usepackage{tabularx}
\usepackage{pgfplots}
\usepackage{tikz}
\usetikzlibrary{
	patterns,
	arrows,
	shapes.misc,
	shapes.symbols,
	calc,
	backgrounds,
	positioning,
}

\makeatletter
\def\pgfmathfloatpow@#1#2{%
    \begingroup%
    \expandafter\pgfmathfloat@decompose@tok#1\relax\pgfmathfloat@a@S\pgfmathfloat@a@Mtok\pgfmathfloat@a@E
    \ifcase\pgfmathfloat@a@S\relax
        \pgfmathfloatcreate{0}{0.0}{0}%
    \else
        \expandafter\pgfmathfloat@decompose@tok#2\relax\pgfmathfloat@a@S\pgfmathfloat@a@Mtok\pgfmathfloat@a@E
        \ifcase\pgfmathfloat@a@S\relax
            \pgfmathfloatcreate{1}{1.0}{0}%
        \or
            \pgfmathfloatpow@@{#1}{#2}%
        \or
            \pgfmathfloatpow@@{#1}{#2}%
        \or
            \edef\pgfmathresult{#2}%
        \or
            \edef\pgfmathresult{#2}%
        \or
            \pgfmathfloatcreate{0}{0.0}{0}%
        \fi
    \fi
    \pgfmath@smuggleone\pgfmathresult
    \endgroup
}%
\makeatother

\newcolumntype{Y}{>{\centering\arraybackslash}X}

\newcommand{\bx}{\mathbf{x}}
\newcommand{\bz}{\mathbf{z}}
\newcommand{\by}{\mathbf{y}}

\definecolor{frrn_col}         {RGB}{  0,255,  0}
\definecolor{lrr_col}          {RGB}{255,127,  0}
\definecolor{dilation_col}     {RGB}{255,  0,  0}
\definecolor{enet_col}         {RGB}{220,220,  0}
\definecolor{adelaide_col}     {RGB}{244, 35,232}
\definecolor{segnet_col}       {RGB}{  0,  0,255}
\definecolor{deeplab_col}      {RGB}{  0,255,255}
\definecolor{gray_col}	       {RGB}{180,180,180}

\definecolor{res_blue}         {RGB}{  0,  0,255}
\definecolor{pool_red}         {RGB}{255,  0,  0}

\definecolor{csunlabeled}     {RGB}{  0,  0,  0}
\definecolor{csroad}          {RGB}{128, 64,128}
\definecolor{cssidewalk}      {RGB}{244, 35,232}
\definecolor{csbuilding}      {RGB}{ 70, 70, 70}
\definecolor{cswall}          {RGB}{102,102,156}
\definecolor{csfence}         {RGB}{190,153,153}
\definecolor{cspole}          {RGB}{153,153,153}
\definecolor{cstrafficlight}  {RGB}{250,170, 30}
\definecolor{cstrafficsign}   {RGB}{220,220,  0}
\definecolor{csvegetation}    {RGB}{107,142, 35}
\definecolor{csterrain}       {RGB}{152,251,152}
\definecolor{cssky}           {RGB}{ 70,130,180}
\definecolor{csperson}        {RGB}{220, 20, 60}
\definecolor{csrider}         {RGB}{255,  0,  0}
\definecolor{cscar}           {RGB}{  0,  0,142}
\definecolor{cstruck}         {RGB}{  0,  0, 70}
\definecolor{csbus}           {RGB}{  0, 60,100}
\definecolor{cstrain}         {RGB}{  0, 80,100}
\definecolor{csmotorcycle}    {RGB}{  0,  0,230}
\definecolor{csbicycle}       {RGB}{119, 11, 32}

\newcommand{\imsize}{small_}
\pgfplotsset{
	discard if not/.style 2 args={
		x filter/.code={
			\edef\tempa{\thisrow{#1}}
			\edef\tempb{#2}
			\ifx\tempa\tempb
			\else
				\def\pgfmathresult{inf}
			\fi
		}
	}
}
\pgfplotsset{
	discard if/.style 2 args={
		x filter/.code={
			\edef\tempa{\thisrow{#1}}
			\edef\tempb{#2}
			\ifx\tempa\tempb
				\def\pgfmathresult{inf}
			\else
			\fi
		}
	}
}
\pgfplotsset{
	discard if not and/.style n args={4}{
		x filter/.code={
		\edef\tempa{\thisrow{#1}}
		\edef\tempb{#2}
		\edef\tempc{\thisrow{#3}}
		\edef\tempd{#4}
		\ifx\tempa\tempb
			\ifx\tempc\tempd
			\else
				\def\pgfmathresult{inf}
			\fi
		\else
			\def\pgfmathresult{inf}
		\fi
		}
	}
}

\usepackage{makecell}
\usepackage{ifthen}
\usepackage{stackengine}
\newcommand{\ioucirc}{\begin{Large}
\topinset{\textcolor{black}{$\circ$}}{\textcolor{black}{$\circ$}}{2.5pt}{}
\end{Large}}
\newcommand{\mycirc}[1][black]{\begin{tikzpicture}[baseline=-0.75ex]
\draw [fill=#1, #1] (0,0) circle (2.5pt);
\draw[transparent] (0,0) -- (-7pt, 0);
\draw[transparent] (0,0) -- (4.8pt, 0);
\end{tikzpicture} }

\usepackage[pagebackref=true,breaklinks=true,letterpaper=true,colorlinks,bookmarks=false]{hyperref}

\cvprfinalcopy % *** Uncomment this line for the final submission

\def\cvprPaperID{1691} % *** Enter the CVPR Paper ID here

\newcommand{\PAR}[1]{\vskip4pt \noindent {\bf #1~}}
\newcommand{\PARbegin}[1]{\noindent {\bf #1~}}
\begin{document}

%%%%%%%%% TITLE
\title{Full-Resolution Residual Networks for Semantic Segmentation in Street Scenes}

\author{Tobias Pohlen \hspace{10mm} Alexander Hermans \hspace{10mm} Markus Mathias \hspace{10mm} Bastian Leibe\\
Visual Computing Institute\\
RWTH Aachen University\\
{\tt\small tobias.pohlen@rwth-aachen.de, \{hermans, mathias, leibe\}@vision.rwth-aachen.de}
% For a paper whose authors are all at the same institution,
% omit the following lines up until the closing ``}''.
% Additional authors and addresses can be added with ``\and'',
% just like the second author.
% To save space, use either the email address or home page, not both
}

\maketitle

\input{nn_styles}
%%%%%%%%% ABSTRACT
\begin{abstract}
Semantic image segmentation is an essential component of modern autonomous driving systems, as
an accurate understanding of the surrounding scene is crucial to navigation and action planning.
Current state-of-the-art approaches in semantic image segmentation rely on pre-trained networks that were initially developed for classifying images as a whole.
While these networks exhibit outstanding recognition performance (\ie, what is visible?), they lack localization accuracy (\ie, where precisely is something located?).
Therefore, additional processing steps have to be performed in order to obtain pixel-accurate segmentation masks at the full image resolution.
To alleviate this problem we propose a novel ResNet-like architecture that exhibits strong localization and recognition performance.
We combine multi-scale context with pixel-level accuracy by using two processing streams within our network:
One stream carries information at the full image resolution, enabling precise adherence to segment boundaries.
The other stream undergoes a sequence of pooling operations to obtain robust features for recognition.
The two streams are coupled at the full image resolution using residuals.
Without additional processing steps and without pre-training, our approach achieves an intersection-over-union score of 71.8\% on the Cityscapes dataset.
\vspace{-5pt}
\end{abstract}

%%%%%%%%% BODY TEXT
\vspace{-10pt}
\section{Introduction}

Recent years have seen an increasing interest in self driving cars and in driver assistance systems.
A crucial aspect of autonomous driving is to acquire a comprehensive understanding of the surroundings in which a car is moving.
Semantic image segmentation~\cite{Shotton08CVPR, Long15CVPR, Gould08IJCV, xiao09ICCV, Ladicky14CVPR}, the task of assigning a set of predefined class labels to image pixels, is an important tool for modeling the complex relationships of the semantic entities usually found in street scenes, such as cars, pedestrians, road, or sidewalks.
In automotive scenarios it is used in various ways, \eg as a pre-processing step to discard image regions that are unlikely to contain objects of interest~\cite{Osep16ICRA, Ess2009BMVC}, to improve object detection~\cite{Bansal09ICCVW, Gu09CVPR, Hariharan14ECCV, Zhu15CVPR}, or in combination with 3D scene geometry~\cite{Kundu14jECCV, Floros2012CVPR, Liu10CVPR}.
Many of those applications require precise region boundaries \cite{Ghiasi16ECCV}. In this work, we therefore pursue the goal of achieving high-quality semantic segmentation with precise boundary adherence.

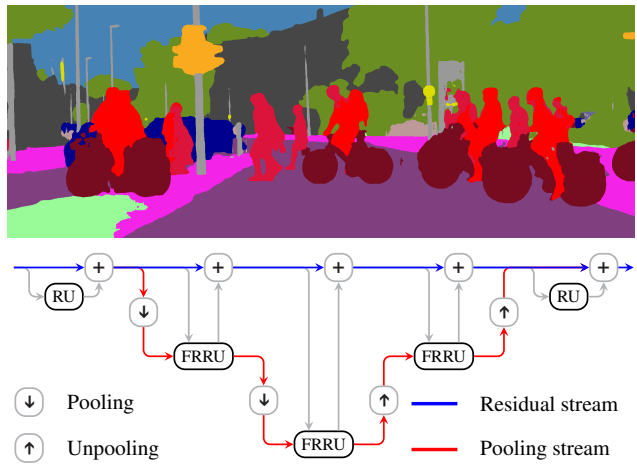
\begin{figure}[t]
\centering
\includegraphics[width=\linewidth]{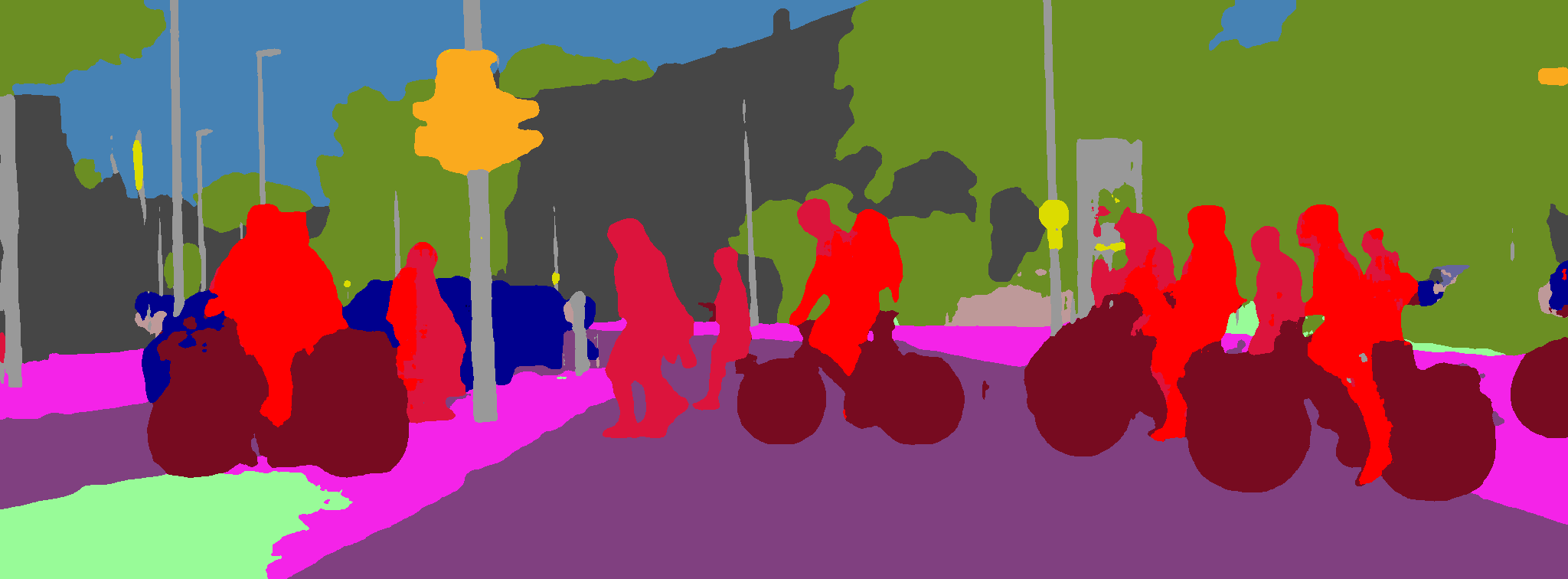}%
\vspace*{5pt}
\resizebox{\linewidth}{!}{
\begin{tikzpicture}

\coordinate (p0) {};

\coordinate[left=0.25cm of p0] (m0) {};

\coordinate[left=0.25cm of m0] (q0);

%\node[inner sep=0pt, left=0.1cm of q0] (input) {\includegraphics[width=1.8cm]{images/results/ims/\imsize munster_000167_000019}};

\node [nn-node, below right=0.25cm and 0.25cm of m0] (p1) {\footnotesize RU};
\node [nn-node, nn-minor] (p2) at ([xshift=0.25cm] p1.north east |- q0) {\textbf{+}};

\node [nn-node, nn-minor, below right=0.25cm and 0.25cm of p2] (p3) {\rotatebox[origin=c]{-90}{\MVRightarrow}};
\node [nn-node, below right=0.25cm and 0.25cm of p3] (p4) {\footnotesize FRRU};

\node [nn-node, nn-minor] (p5) at ([xshift=-0.25cm] p4.north east |- q0) {\textbf{+}};

\node [nn-node, nn-minor, below right=0.25cm and 0.25cm of p4] (p6) {\rotatebox[origin=c]{-90}{\MVRightarrow}};
\node [nn-node, below right=0.25cm and 0.25cm of p6] (p7) {\footnotesize FRRU};

\node [nn-node, nn-minor] (p8) at ([xshift=-0.25cm] p7.north east |- q0) {\textbf{+}};

\node [nn-node, nn-minor, above right=0.25cm and 0.25cm of p7] (p9) {\hspace{0.0425cm}\rotatebox[origin=c]{90}{\MVRightarrow}};
\node [nn-node, above right=0.25cm and 0.25cm of p9] (p10) {\footnotesize FRRU};

\node [nn-node, nn-minor] (p11) at ([xshift=-0.25cm] p10.north east |- q0) {\textbf{+}};

\node [nn-node, nn-minor, above right=0.25cm and 0.25cm of p10] (p12) {\hspace{0.0425cm}\rotatebox[origin=c]{90}{\MVRightarrow }};

\coordinate (p13) at ([xshift=0.25cm] p12.north east |- q0);
\node [nn-node, below right=0.25cm and 0.25cm of p13] (p14) {\footnotesize RU};
\node [nn-node, nn-minor] (p15) at ([xshift=0.25cm] p14.north east |- q0) {\textbf{+}};

\coordinate[right=0.25cm of p15] (p16);

%\node[inner sep=0pt, right=0.1cm of p16] (output) {\includegraphics[width=1.8cm]{images/results/ours/\imsize munster_000167_000019}};

\node [nn-node, nn-minor] (v0) at ([yshift=-0.75cm] m0 |- p4) {\rotatebox[origin=c]{-90}{\MVRightarrow}};
\node [nn-node, nn-minor] (v1) at ([yshift=-0.75cm] m0 |- v0) {\hspace{0.0425cm}\rotatebox[origin=c]{90}{\MVRightarrow}};

\node [anchor=west, minimum height=0.45cm] (v2) at ([xshift=0.25cm, yshift=-0.025cm] p0 |- v0) {Pooling};
\node [anchor=west, minimum height=0.45cm] (v3) at ([xshift=0.25cm, yshift=-0.025cm] p0 |- v1) {Unpooling};

\draw[nn-edge, res_blue, ultra thick] ([xshift=6.0cm, yshift=-0.025cm] p0 |- v0) -- ([xshift=6.75cm, yshift=-0.025cm] p0 |- v0);
\draw[nn-edge, pool_red, ultra thick] ([xshift=6.0cm, yshift=-0.025cm] p0 |- v1) -- ([xshift=6.75cm, yshift=-0.025cm] p0 |- v1);

\node [anchor=west, minimum height=0.45cm] (v4) at ([xshift=7.0cm, yshift=-0.025cm] p0 |- v0) {Residual stream};
\node [anchor=west, minimum height=0.45cm] (v5) at ([xshift=7.0cm, yshift=-0.025cm] p0 |- v1) {Pooling stream};

\draw[nn-edge, nn-arrow, Gray!70] (q0) -- (m0) |- (p1);
\draw[nn-edge, nn-arrow, res_blue] (q0) -- (p2);
\draw[nn-edge, nn-arrow, Gray!70] (p1) -| (p2);
\draw[nn-edge, nn-arrow, pool_red] (p2) -| (p3);
\draw[nn-edge, nn-arrow, pool_red] (p3) |- (p4);
\draw[nn-edge, nn-arrow, Gray!70] (p2) -| ([xshift=0.25cm] p4.north west);
\draw[nn-edge, nn-arrow, Gray!70] ([xshift=-0.25cm] p4.north east) -- (p5);
\draw[nn-edge, nn-arrow, res_blue] (p2) -- (p5);

\draw[nn-edge, nn-arrow, pool_red] (p4) -| (p6);
\draw[nn-edge, nn-arrow, pool_red] (p6) |- (p7);

\draw[nn-edge, nn-arrow, Gray!70] (p5) -| ([xshift=0.25cm] p7.north west);
\draw[nn-edge, nn-arrow, Gray!70] ([xshift=-0.25cm] p7.north east) -- (p8);
\draw[nn-edge, nn-arrow, res_blue] (p5) -- (p8);

\draw[nn-edge, nn-arrow, pool_red] (p7) -| (p9);
\draw[nn-edge, nn-arrow, pool_red] (p9) |- (p10);

\draw[nn-edge, nn-arrow, Gray!70] (p8) -| ([xshift=0.25cm] p10.north west);
\draw[nn-edge, nn-arrow, Gray!70] ([xshift=-0.25cm] p10.north east) -- (p11);
\draw[nn-edge, nn-arrow, res_blue] (p8) -- (p11);

\draw[nn-edge, nn-arrow, pool_red] (p10) -| (p12);

\draw[nn-edge, nn-arrow, pool_red] (p12) |- (p15);

\draw[nn-edge, nn-arrow, Gray!70] (p14) -| (p15);
\draw[nn-edge, nn-arrow, Gray!70] (p11) -- (p13) |- (p14);
\draw[nn-edge, nn-arrow, res_blue] (p15) -? (p16);
\draw[nn-edge, nn-arrow, res_blue] (p11) -- (p15);

\end{tikzpicture}}
\caption{
Example output and the abstract structure of our full-resolution residual network.
The network has two processing streams.
The residual stream (blue) stays at the full image resolution, the pooling stream (red) undergoes a sequence of pooling and unpooling operations.
The two processing streams are coupled using full-resolution residual units (FRRUs).
}
\label{fig:overview}
\vspace{-3pt}
\end{figure}

Current state-of-the-art approaches for image segmentation all employ some form of \emph{fully convolutional network (FCNs)} \cite{Long15CVPR} that takes the image as input and outputs a probability map for each class. Many papers rely on network architectures that have already been proven successful for image classification such as variants of the ResNet \cite{He16CVPR} or the VGG architecture \cite{Simonyan15ICLR}.
Starting from pre-trained nets, where a large number of weights for the target task can be pre-set by an auxiliary classification task, reduces training time and often yields superior performance compared to training a network from scratch using the (possibly limited amount of) data of the target application.
%(BL)
However, a main limitation of using such pre-trained networks is that they severely restrict the design space of novel approaches,
since new network elements such as batch normalization \cite{Ioffe15ICML} or new activation functions often cannot be added into an existing architecture.

When performing semantic segmentation using FCNs, a common strategy is to successively reduce the spatial size of the feature maps using pooling operations or strided convolutions.
This is done for two reasons: First, it significantly increases the size of the receptive field and second, it makes the network robust against small translations in the image.
While pooling operations are highly desirable for recognizing objects in images, they significantly deteriorate localization performance of the networks when applied to semantic image segmentation.
Several approaches exist to overcome this problem and obtain pixel-accurate segmentations.
Noh \etal~\cite{Noh15ICCV} learn a mirrored VGG network as a decoder, Yu and Koltun~\cite{Yu16ICLR} introduce dilated convolutions to reduce the pooling factor of their pre-trained network. Ghiasi \etal~\cite{Ghiasi16ECCV} use multi-scale predictions to successively improve their boundary adherence.
An alternative approach used by several methods is to apply post-processing steps such as CRF-smoothing \cite{Krahenbuhl11NIPS}.

In this paper, we propose a novel network architecture that achieves state-of-the-art segmentation performance without the need for additional post-processing steps and without the limitations imposed by pre-trained architectures.
%, in contrast to the vast majority of today's semantic segmentation approaches.
Our proposed ResNet-like architecture unites strong recognition performance with precise localization capabilities by combining two distinct processing streams.
%Our novel ResNet-like network architecture combines strong recognition performance with strong localization performance by using two processing streams: 
One stream undergoes a sequence
of pooling operations and is responsible for understanding large-scale relationships of image elements; the other stream carries feature maps at the full image resolution, resulting in precise boundary adherence.
This idea is visualized in Figure \ref{fig:overview}, where the two processing streams are shown in blue and red. 
The blue residual lane reflects the high-resolution stream.
It can be combined with classical residual units (left and right), as well as with our new full-resolution residual units (FRRU).
The FRRUs from the red pooling lane act as residual units for the blue stream, but also undergo pooling operations and carry high-level information through the network.
This results in a network that successively combines and computes features at two resolutions.

This paper makes the following contributions: 
\textit{(i)} We propose a novel network architecture geared towards precise semantic segmentation in street scenes which is not limited to pre-trained architectures and achieves state-of-the-art results.
\textit{(ii)} We propose to use two processing streams to realize strong recognition and strong localization performance:
One stream undergoes a sequence of pooling operations while the other stream stays at the full image resolution.
\textit{(iii)} In order to foster further research in this area, we publish our code and the trained models in Theano/Lasagne~\cite{Theano16, Lasagne15}\footnote{https://github.com/TobyPDE/FRRN}.

\vspace{-5pt}
\section{Related Work}

The dramatic performance improvements from using CNNs for semantic segmentation have brought about an increasing demand for such algorithms in the context of autonomous driving scenarios.
As a large amount of annotated data is crucial in order to train such deep networks, multiple new datasets have been released to encourage further research in this area, including Synthia~\cite{Ros16CVPR}, Virtual KITTI~\cite{Gaidon16CVPR}, and Cityscapes~\cite{Cordts16CVPR}.
In this work, we focus on Cityscapes, a recent large-scale dataset consisting of real-world imagery with well-curated annotations. Given their success, we will constrain our literature review to deep learning based semantic segmentation approaches and deep learning network architectures.

\PAR{Semantic Segmentation Approaches.}
Over the last years, the most successful semantic segmentation approaches have been based on convolutional neural networks (CNNs).
Early approaches constrained their output to a bottom-up segmentation followed by a CNN based region classification~\cite{yan15CVPR}.
Rather than classifying entire regions in the first place, the approach by Farabet \etal performs pixel-wise classification using CNN features originating from multiple scales, followed by aggregation of these noisy pixel predictions over superpixel regions \cite{Farabet13TPAMI}.

The introduction of so-called \emph{fully convolutional networks (FCNs)} for semantic image segmentation by Long \etal \cite{Long15CVPR} opened a wide range of semantic segmentation research using end-to-end training~\cite{Dai2015CVPR}.
Long \etal further reformulated the popular VGG architecture \cite{Simonyan15ICLR} as a fully convolutional network (FCN), enabling the use of pre-trained models for this architecture.
To improve segmentation performance at object boundaries, \emph{skip connections} were added which allow information to propagate directly from early, high-resolution layers to deeper layers.

Pooling layers in FCNs fulfill a crucial role in order to increase the receptive field size of later units and with it the classification performance. However, they have the downside that the resulting network outputs are at a lower resolution.
To overcome this, various strategies have been proposed.
Some approaches extract features from intermediate layers via some sort of skip connections \cite{Long15CVPR, Chen15ICLR, Liu2015ICLRW, HChen16ARXIV}.
Noh \etal propose an encoder/decoder network \cite{Noh15ICCV}.
The encoder computes low-dimensional feature representations via a sequence of pooling and convolution operations.
The decoder, which is stacked on top of the encoder, then learns an upscaling of these low-dimensional features via subsequent unpooling and deconvolution operations \cite{Zeiler11CVPR}.
Similarly, Badrinarayanan \etal \cite{Badrinarayanan15aARXIV, Badrinarayanan15bARXIV} use convolutions instead of deconvolutions in the decoder network.
%(BL)
In contrast, our approach preserves high-resolution information throughout the entire network by keeping a separate high-resolution processing stream.
%>our approach two processing streams.

Many approaches apply smoothing operations to the output of a CNN in order to obtain more consistent predictions.
Most commonly, \emph{conditional random fields (CRFs)} \cite{Krahenbuhl11NIPS} are applied on the network output \cite{Chen16ARXIV, Chen15ICLR, Dai15ICCV, Lin16CVPR, Chandra16ARXIV}.
%(BL)
More recently, some papers approximate the \emph{mean-field inference} of CRFs using specialized network architectures \cite{Zheng15ICCV, Schwing15ARXIV, Liu15ICCV}.
Other approaches to smoothing the network predictions include \emph{domain transform} \cite{Chen15ICLR, Gastal11TOG} and superpixel-based smoothing \cite{Farabet13TPAMI, Mostajabi15CVPR}.
Our approach is able to swiftly combine high- and low-resolution information, resulting in already smooth output predictions.
Experiments with additional CRF smoothing therefore did not result in significant performance improvements.

\PAR{Network architectures.}
Since the success of the AlexNet architecture \cite{Krizhevsky12NIPS} in the ImageNet Large-Scale Visual Classification Challenge (ILSVRC) \cite{Russakovsky15IJSCV}, the vision community has seen several milestones with respect to CNN architectures. The network depth has been constantly increased, first with the popular VGG net \cite{Simonyan15ICLR}, then by using batch normalization with GoogleNet \cite{Szegedy15CVPR}. Lately, many computer vision applications have adopted the ResNet architecture \cite{He16CVPR}, which often leads to signification performance boosts compared to earlier network architectures.
All of these developments show how important a proper architecture is. However, so far most of these networks have been specifically tailored towards the task of classification,
%(BL)
in many cases including a pre-training step on ILSVRC. As a result, some of their design choices may contribute to a suboptimal performance when performing pixel-to-pixel tasks such as semantic segmentation.
In contrast, our proposed architecture has been specifically designed for segmentation tasks
and reaches competitive performance on the Cityscapes dataset without requiring ILSVRC pre-training.
\section{Network Architectures for Segmentation}

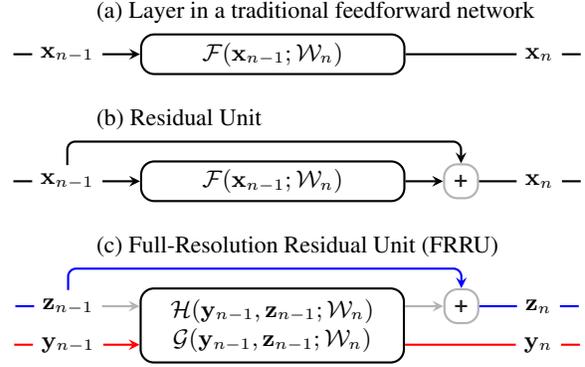
\begin{figure}[!]

%\begin{tabularx}{\linewidth}{@{}Y@{}}
	\hspace*{1.5cm}\small (a) Layer in a traditional feedforward network\\
	\vspace{-0.5cm}
	\begin{center}
	\begin{tikzpicture}
		\node (p0) {$\bx_{n-1}$};
		\node[right=0.5cm of p0, nn-node, minimum width=3.5cm] (p1) {$\mathcal{F}(\bx_{n-1}; \mathcal{W}_n)$};
		\node[right=1.4775cm of p1] (p3) {$\bx_{n}$};
		
		\coordinate[left=0.25cm of p0] (q0);
		\coordinate[right=0.25cm of p3] (q1);
		\draw[nn-edge] (q0) -- (p0);
		\draw[nn-edge, nn-arrow] (p0) -- (p1);
		\draw[nn-edge] (p3) -- (q1);
		\draw[nn-edge] (p1) -- (p3);
	\end{tikzpicture}
	\end{center}
	\hspace*{1.5cm}\small (b) Residual Unit\\
	\vspace{-0.5cm}
	\begin{center}
	\begin{tikzpicture}
		\node (p0) {$\bx_{n-1}$};
		\node[right=0.5cm of p0, nn-node, minimum width=3.5cm] (p1) {$\mathcal{F}(\bx_{n-1}; \mathcal{W}_n)$};
		\node[right=0.5cm of p1, nn-node, nn-minor] (p2) {\textbf{+}};
		\node[right=0.5cm of p2] (p3) {$\bx_{n}$};
		
		\coordinate[left=0.25cm of p0] (q0);
		\coordinate[right=0.25cm of p3] (q1);
		\coordinate[above=0.25cm of p1] (q2);
		\draw[nn-edge] (q0) -- (p0);
		\draw[nn-edge, nn-arrow] (p0) -- (p1);
		\draw[nn-edge, nn-arrow] (p1) -- (p2);
		\draw[nn-edge, nn-arrow] (p0) |- (q2) -| (p2);
		\draw[nn-edge] (p2) -- (p3);
		\draw[nn-edge] (p3) -- (q1);
	\end{tikzpicture}
	\end{center}
	\hspace*{1.5cm}\small (c) Full-Resolution Residual Unit (FRRU)\\
	\vspace*{-0.5cm}
	\begin{center}
	\begin{tikzpicture}
		\node[nn-node, minimum width=3.5cm] (p1) {
			\begin{tabular}{c}
			$\mathcal{H}(\by_{n-1}, \bz_{n-1};\mathcal{W}_n)$ \\ $\mathcal{G}(\by_{n-1}, \bz_{n-1}; \mathcal{W}_n)$
			\end{tabular}
		};
		\node (p0) at ([xshift=-0.95cm, yshift=-0.25cm] p1.north west) {$\bz_{n-1}$};
		\node (v0) at ([xshift=-0.95cm, yshift=0.25cm] p1.south west) {$\by_{n-1}$};
		
		\node[nn-node, nn-minor] (p2) at ([xshift=0.75cm, yshift=-0.25cm] p1.north east) {\textbf{+}};
		\node[right=0.5cm of p2] (p3) {$\bz_{n}$};
		\node (v1) at ([yshift=0.25cm] p3 |- p1.south west) {$\by_{n}$};
		\coordinate[above=0.25cm of p1] (q2);
		\coordinate[left=0.25cm of p0] (q0);
		\coordinate[right=0.25cm of p3] (q1);

		\coordinate[left=0.25cm of v0] (u0);
		\coordinate[right=0.25cm of v1] (u1);
		
		\draw[nn-edge, nn-arrow, res_blue] (p0) |- (q2) -| (p2);		
		\draw[nn-edge, res_blue] (q0) -- (p0);
		
		\draw[nn-edge, nn-arrow, Gray!70] (p0) -- ([yshift=-0.25cm] p1.north west);
		\draw[nn-edge, nn-arrow, Gray!70] ([yshift=-0.25cm] p1.north east) -- (p2);
		
		\draw[nn-edge, res_blue] (p2) -- (p3);
		\draw[nn-edge, res_blue] (p3) -- (q1);
						
		\draw[nn-edge, nn-arrow, pool_red] (v0) -- ([yshift=0.25cm] p1.south west);
		\draw[nn-edge, pool_red] ([yshift=0.25cm] p1.south east) -- (v1);
		\draw[nn-edge, pool_red] (u1) -- (v1);
		\draw[nn-edge, pool_red] (u0) -- (v0);
	\end{tikzpicture}
	\end{center}
	
	\caption{The figure compares the structures of different network design elements. (a) shows a layer in a traditional feedforward network; (b) shows a residual unit; (c) shows a full-resolution residual unit.}
	\label{fig:compare}
%\end{tabularx}

\end{figure}

\PARbegin{Feed-Forward Networks.}
Until recently, the majority of feedforward networks, such as the VGG-variants \cite{Simonyan15ICLR}, were composed of a linear sequence of layers.
Each layer in such a network computes a function $\mathcal{F}$ and the output $\bx_{n}$ of the $n$-th layer is computed as
\begin{align}
\bx_{n} = \mathcal{F}(\bx_{n-1}; \mathcal{W}_n)
\end{align}
where $\mathcal{W}_n$ are the parameters of the layer (see \ref{fig:compare}a).
We refer to this class of network architectures as \emph{traditional feedforward networks}.

\PAR{Residual Networks (ResNets).}
He \etal observed that deepening traditional feedforward networks often results in an increased training loss \cite{He16CVPR}.
In theory, however, the training loss of a shallow network should be an upper bound on the training loss of a corresponding deep network.
This is due to the fact that increasing the depth by adding layers strictly increases the expressive power of the model.
A deep network can express all functions that the original shallow network can express by using identity mappings for the added layers.
Hence a deep network should perform at least as well as the shallower model on the training data.
The violation of this principle implied that current training algorithms have difficulties optimizing very deep traditional feedforward networks.
He \etal proposed \emph{residual networks (ResNets)} that exhibit significantly improved training characteristics, allowing network depths that were previously unattainable.

A ResNet is composed of a sequence of \emph{residual units (RUs)}.
As depicted in Figure~\ref{fig:compare}b, the output $\bx_{n}$ of the $n$-th RU in a ResNet is computed as
\begin{align}
	\bx_{n} = \bx_{n - 1} + \mathcal{F}(\bx_{n-1}; \mathcal{W}_n) \label{eq:resnet}
\end{align}
where $\mathcal{F}( \bx_{n-1} \ ; \mathcal{W}_n)$ is the residual, which is parametrized by $\mathcal{W}_n$.
Thus, instead of computing the output $\bx_{n}$ directly, $\mathcal{F}$ only computes a residual that is added to the input $\bx_{n-1}$.
One commonly refers to this design as \emph{skip connection}, because there is a connection from the input $\bx_{n-1}$ to the output $\bx_{n}$ that skips the actual computation $\mathcal{F}$.

It has been empirically observed that ResNets have superior training properties over traditional feedforward networks.
This can be explained by an improved gradient flow within the network.
In oder to understand this, consider the $n$-th and $m$-th residual units in a ResNet where $m > n$ (\ie, the $m$-th unit is closer to the output layer of the network).
By applying the recursion (\ref{eq:resnet}) several times, He \etal showed in  \cite{He16ECCV} that the output of the $m$-th residual unit admits a representation of the form
\begin{align}
	\bx_m = \bx_n + \sum_{i = n}^{m - 1} \mathcal{F}(\bx_i; \mathcal{W}_{i+1}).
\end{align}
Furthermore, if $l$ is the loss that is used to train the network, we can use the chain rule of calculus and express the derivative of the loss $l$ with respect to the output $\bx_n$ of the $n$-th RU as
\begin{align}
	\frac{\partial l}{\partial \bx_n} &= \frac{\partial l}{\partial \bx_m} \frac{\partial \bx_m}{\partial \bx_n} = \frac{\partial l}{\partial \bx_m} + \frac{\partial l}{\partial \bx_m} \sum_{i = n}^{m - 1} \frac{\partial \mathcal{F}(\bx_i; \mathcal{W}_{i+1})}{\partial \bx_n}.
\end{align}
Thus, we find
\begin{align}
	\frac{\partial l}{\partial \mathcal{W}_{n}} &= \frac{\partial l}{\partial \bx_n} \frac{\partial \bx_n}{\partial \mathcal{W}_{n}} \notag \\
	&= \frac{\partial \bx_n}{\partial \mathcal{W}_{n}} \left( \frac{\partial l}{\partial \bx_m} + \frac{\partial l}{\partial \bx_m} \sum_{i = n}^{m - 1} \frac{\partial \mathcal{F}(\bx_i; \mathcal{W}_{i+1})}{\partial \bx_n} \right).
\end{align}
We see that the weight updates depend on two sources of information, $\frac{\partial l}{\partial \bx_m}$ and $\frac{\partial l}{\partial \bx_m} \sum_{i = n}^{m - 1} \frac{\partial \mathcal{F}(\bx_i; \mathcal{W}_{i+1})}{\partial \bx_n}$.
While the amount of information that is contained in the latter may depend crucially on the depth $n$, the former allows a gradient flow that is independent of the depth.
Hence, gradients can flow unhindered from the deeper unit to the shallower unit.
This makes training even extremely deep ResNets possible.

\PAR{Full-Resolution Residual Networks (FRRNs).}
In this paper, we unify the two above-mentioned principles of network design and propose \emph{full-resolution residual networks (FRRNs)} that exhibit the same superior training properties as ResNets but have two processing streams.
The features on one stream, the \emph{residual stream}, are computed by adding successive residuals, while the features on the other stream, the \emph{pooling stream}, are the direct result of a sequence of convolution and pooling operations applied to the input.

Our design is motivated by the need to have networks that can jointly compute good high-level features for recognition and good low-level features for localization.
Regardless of the specific network design, obtaining good high-level features requires a sequence of pooling operations.
The pooling operations reduce the size of the feature maps and increase the network's receptive field, as well as its robustness against small translations in the image.
While this is crucial to obtaining robust high-level features, networks that employ a deep pooling hierarchy have difficulties tracking low-level features, such as edges and boundaries, in deeper layers.
This makes them good at recognizing the elements in a scene but bad at localizing them to pixel accuracy.
On the other hand, a network that does not employ any pooling operations behaves the opposite way.
It is good at localizing object boundaries, but performs poorly at recognizing the actual objects.
By using the two processing streams together, we are able to compute both kinds of features simultaneously.
While the residual stream of an FRRN computes successive residuals at the full image resolution, allowing low level features to propagate effortlessly through the network,
the pooling stream undergoes a sequence of pooling and unpooling operations resulting in good high-level features.
Figure \ref{fig:overview} visualizes the concept of having two distinct processing streams.

An FRRN is composed of a sequence of \emph{full-resolution residual units (FRRUs)}.
Each FRRU has two inputs and two outputs, because it simultaneously operates on both streams.
Figure \ref{fig:compare}c shows the structure of an FRRU.
Let $\bz_{n-1}$ be the residual input to the $n$-th FRRU and let $\by_{n-1}$ be its pooling input.
Then the outputs are computed as
\begin{align}
	\bz_{n} & = \bz_{n - 1} + \mathcal{H}(\by_{n-1}, \bz_{n-1}; \mathcal{W}_n)\\
	\by_{n} & = \mathcal{G}(\by_{n-1}, \bz_{n-1}; \mathcal{W}_n),
\end{align}
where $\mathcal{W}_n$ are the parameters of the functions $\mathcal{G}$ and $\mathcal{H}$, respectively.

If ${\mathcal{G}\equiv0}$, then an FRRU corresponds to an RU since it disregards the pooling input $\by_n$, and the network effectively becomes an ordinary ResNet.
On the other hand, if ${\mathcal{H}\equiv0}$, then the output of an FRRU only depends on its input via the function $\mathcal{G}$.
Hence, no residuals are computed and we obtain a traditional feedforward network.
By carefully constructing $\mathcal{G}$ and $\mathcal{H}$, we can combine the two network principles.

In order to show that FRRNs have similar training characteristics as ResNets, we adapt the analysis presented in \cite{He16ECCV} to our case.
Using the same recursive argument as before, we find that for $m > n$, $\bz_m$ has the representation
\begin{align}
	\bz_m = \bz_n + \sum_{i = n}^{m - 1} \mathcal{H}(\by_i, \bz_i; \mathcal{W}_{i+1}).
\end{align}
We can then express the derivative of the loss $l$ with respect to the weights $\mathcal{W}_{n}$ as
\begin{align}
	\frac{\partial l}{\partial \mathcal{W}_{n}} =& \frac{\partial l}{\partial \bz_n}\frac{\partial \bz_n}{\partial \mathcal{W}_{n}} + \frac{\partial l}{\partial \by_n}\frac{\partial \by_n}{\partial \mathcal{W}_{n}} \notag \\
	=&\frac{\partial \bz_n}{\partial \mathcal{W}_{n}} \left( \frac{\partial l}{\partial \bz_m} + \frac{\partial l}{\partial \bz_m} \sum_{i = n}^{m - 1} \frac{\partial \mathcal{H}(\by_i, \bz_i; \mathcal{W}_{i+1})}{\partial \bz_n} \right) \notag \\
	& +\frac{\partial l}{\partial \by_n}\frac{\partial \by_n}{\partial \mathcal{W}_{n}}.
\end{align}
Hence, the weight updates depend on three sources of information.
Analogous to the analysis of ResNets, the two sources $\frac{\partial l}{\partial \by_n}\frac{\partial \by_n}{\partial \mathcal{W}_{n}}$ and $\frac{\partial l}{\partial \bz_m} \sum_{i = n}^{m - 1} \frac{\partial \mathcal{H}(\by_i, \bz_i; \mathcal{W}_{i+1})}{\partial \bz_n}$ depend crucially on the depth $n$, while the term $\frac{\partial l}{\partial \bz_m}$ is independent of the depth.
Thus, we achieve a depth-independent gradient flow for all parameters that are used by the residual function $\mathcal{H}$.
If we use some of these weights in order to compute the output of $\mathcal{G}$, all weights of the unit benefit from the improved gradient flow.
This is most easily achieved by reusing the output of $\mathcal{G}$ in order to compute $\mathcal{H}$.
However, we note that other designs are possible.

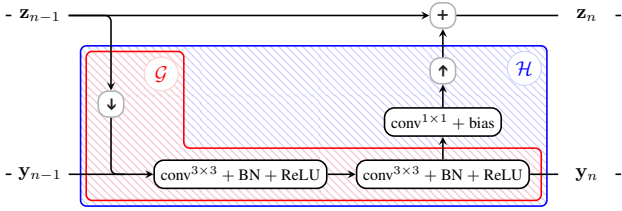
\begin{figure}[t]

\resizebox{\linewidth}{!}{
\begin{tikzpicture}[auto,scale = 1]

\node [minimum width=1cm, minimum height=0.8cm] (v1) {$\bz_{n-1}$};
\node [minimum width=1cm, below=2cm of v1, minimum height=0.8cm] (v2) {$\by_{n-1}$};

\coordinate[left=0.1cm of v1] (c1);
\coordinate[left=0.1cm of v2] (c2);

\coordinate[right=1cm of v2] (c4);
\coordinate[right=0.75cm of v2] (c5);

\node [nn-node, nn-minor, above=1cm of c5, fill=white] (c3) {\rotatebox[origin=c]{-90}{\MVRightarrow}};

\node [nn-node, right=0.5cm of c4, fill=white] (n1) {\footnotesize $\text{conv}^{3 \times 3} + \text{BN} + \text{ReLU}$};
\node [nn-node, right=0.5cm of n1, fill=white] (n2) {\footnotesize $\text{conv}^{3 \times 3} + \text{BN} + \text{ReLU}$};

\node [nn-node, above=0.4cm of n2, fill=white] (n3) {\footnotesize $\text{conv}^{1 \times 1} + \text{bias}$};
\node [nn-node, nn-minor, above=0.4cm of n3, fill=white] (n4) {\hspace{0.0425cm}\rotatebox[origin=c]{90}{\MVRightarrow}};

\node [nn-node, nn-minor, fill=white] (n5) at (n4 |- v1) {\textbf{+}};

\node [minimum width=1cm, right=0.5cm of n2, minimum height=0.8cm] (v4) {$\by_{n}$};
\node [minimum width=1cm, above=2cm of v4, minimum height=0.8cm] (v3) {$\bz_{n}$};

\coordinate[right=0.1cm of v3] (c7);
\coordinate[right=0.1cm of v4] (c8);

\draw[nn-edge] (c1) -- (v1);
\draw[nn-edge] (c2) -- (v2);

\draw[nn-edge, nn-arrow] (v1) -| (c3);
\draw[nn-edge] (c3) |- (c4);
\draw[nn-edge] (v2) -- (c4);
\draw[nn-edge, nn-arrow] (c4) -- (n1);
\draw[nn-edge, nn-arrow] (n1) -- (n2);
\draw[nn-edge, nn-arrow] (n2) -- (n3);
\draw[nn-edge, nn-arrow] (n3) -- (n4);
\draw[nn-edge, nn-arrow] (n4) -- (n5);
\draw[nn-edge, nn-arrow] (v1) -- (n5);
\draw[nn-edge] (n2) -- (v4);
\draw[nn-edge] (n5) -- (v3);
\draw[nn-edge] (v3) -- (c7);
\draw[nn-edge] (v4) -- (c8);

\node[text=red] (t1) at ([xshift=0.65cm, yshift=0.29cm] c3.north east) {$\mathcal{G}$};
\node[text=blue] (t2) at ([xshift=-0.1cm, yshift=-0.2cm] n2.south east |- n4.north east) {$\mathcal{H}$};

\begin{pgfonlayer}{background}

\draw[nn-edge, res_blue, pattern=north west lines, pattern color=res_blue!20] ([xshift=-0.3cm, yshift=0.3] c3.south west |- n4.north) -- ([xshift=-0.3cm, yshift=-0.3cm] c3.north west |- n1.south) -- ([xshift=0.3cm, yshift=-0.3cm] n2.south east) -- ([xshift=0.3cm, yshift=0.2cm] n2.south east |- n4.north east) -| ([xshift=-0.3cm, yshift=0.2] c3.south west |- n4.north east);
\draw[nn-edge, pool_red, pattern=north west lines, pattern color=pool_red!20] ([xshift=-0.2cm] c3.south west) -- ([xshift=-0.2cm, yshift=-0.2cm] c3.north west |- n1.south) -- ([xshift=0.2cm, yshift=-0.2cm] n2.south east) -- ([xshift=0.2cm, yshift=0.2cm] n2.north east) -| ([xshift=1.05cm, yshift=0.69cm] c3.north east) -| ([xshift=-0.2cm] c3.south west);
\draw[red!20, fill=white] (t1) circle (0.3cm);
\draw[blue!20, fill=white] (t2) circle (0.3cm);

\end{pgfonlayer}

\end{tikzpicture}
}
\caption{The figure shows our design of a full-resolution residual unit (FRRU).
The inner red box marks the parts of the unit that are computed by the function $\mathcal{G}$ while the outer blue box indicates the parts that are computed by the function $\mathcal{H}$.
}
\label{fig:frru}
\end{figure}

Figure \ref{fig:frru} shows our proposed FRRU design.
The unit first concatenates the two incoming streams by using a pooling layer in order to reduce the size of the residual stream.
Then the concatenated features are fed through two convolution units.
Each convolution unit consists of a $3\times3$ convolution layer followed by a batch normalization layer \cite{Ioffe15ICML} and a ReLU activation function.
The result of the second convolution unit is used in two ways.
First, it forms the \emph{pooling stream} input of the next FRRU in the network and second it is the basis for the computed residual.
To this end, we first adjust the number of feature channels using a $1 \times 1$ convolution and then upscale the spatial dimensions using an unpooling layer.
Because the features might have to be upscaled significantly (\eg, by a factor of 16), we found that simply upscaling by repeating the entries along the spatial dimensions performed superior to bilinear interpolation.

In Figure \ref{fig:frru}, the inner red box corresponds to the function $\mathcal{G}$ while the outer blue box corresponds to the function $\mathcal{H}$.
We can see that the output of $\mathcal{G}$ is used in order to compute $\mathcal{H}$, because the red box is entirely contained within the blue box.
As shown above, this design choice results in superior gradient flow properties for all weights of the unit.

\begin{table}[t]
	\caption{The table shows our two network designs.
	By $\text{conv}_m^{k \times k}$ we denote a convolution layer having $m$ kernels each of size $k \times k$.
	The notations $\text{RU}_m$ and $\text{FRRU}_m$ refer to residual units and full-resolution residual units whose convolutions have $m$ channels, respectively.
	The parameter $c$ indicates the number of classes to predict.
	}
	\footnotesize
	\hspace*{-6pt}\begin{tabular}{cc}
	\begin{tabularx}{0.48\linewidth}{!{\color{pool_red}\vrule width 2pt}Y!{\color{pool_red}\vrule width 2pt}!{\color{res_blue}\vrule width 2pt}Y!{\color{res_blue}\vrule width 2pt}}
		\Xhline{3\arrayrulewidth}
		\multicolumn{2}{c}{ FRRN A} \\
		\hline
		\multicolumn{2}{c}{}\\[-8pt]
		\multicolumn{2}{c}{ $ \text{conv}_{48}^{5 \times 5} + \text{BN} + \text{ReLU} $ } \\[1pt]
		\hline
		\multicolumn{2}{c}{ $ 3 \times \text{RU}_{48}$ } \\
		\cellcolor{pool_red}\textcolor{white}{\footnotesize \textbf{pooling stream}} & \cellcolor{res_blue}\textcolor{white}{\footnotesize \textbf{residual stream}} \\
		& \\[-8pt]
		max pool & $\text{conv}_{32}^{1 \times 1}$ \\[1pt]
		\hline
		\multicolumn{2}{!{\color{pool_red}\vrule width 2pt}c!{\color{res_blue}\vrule width 2pt}}{ $3 \times \text{FRRU}_{96}$ } \\
		\hline
		max pool &  \\
		\hline
		\multicolumn{2}{!{\color{pool_red}\vrule width 2pt}c!{\color{res_blue}\vrule width 2pt}}{ $4 \times \text{FRRU}_{192}$ } \\
		\hline
		max pool &  \\
		\hline
		\multicolumn{2}{!{\color{pool_red}\vrule width 2pt}c!{\color{res_blue}\vrule width 2pt}}{ $2 \times \text{FRRU}_{384}$ } \\
		\hline
		max pool &  \\
		\hline
		\multicolumn{2}{!{\color{pool_red}\vrule width 2pt}c!{\color{res_blue}\vrule width 2pt}}{ $2 \times \text{FRRU}_{384}$ } \\
		\hline
		& \\[3\arrayrulewidth] %fixing the missing height of the hlines.
		& \\
		& \\
		& \\
		\hline
		unpool &  \\
		\hline
		\multicolumn{2}{!{\color{pool_red}\vrule width 2pt}c!{\color{res_blue}\vrule width 2pt}}{ $2 \times \text{FRRU}_{192}$ } \\
		\hline
		unpool &  \\
		\hline
		\multicolumn{2}{!{\color{pool_red}\vrule width 2pt}c!{\color{res_blue}\vrule width 2pt}}{ $2 \times \text{FRRU}_{192}$ } \\
		\hline
		unpool &  \\
		\hline
		\multicolumn{2}{!{\color{pool_red}\vrule width 2pt}c!{\color{res_blue}\vrule width 2pt}}{ $2 \times \text{FRRU}_{96}$ } \\
		\hline
		unpool &  \\
		\cellcolor{pool_red}\textcolor{white}{\footnotesize \textbf{pooling stream}} & \cellcolor{res_blue}\textcolor{white}{\footnotesize \textbf{residual stream}} \\
		\multicolumn{2}{c}{ concatenate } \\
		\hline
		\multicolumn{2}{c}{ $3 \times \text{RU}_{48}$ } \\
		\hline
		\multicolumn{2}{c}{}\\[-8pt]
		\multicolumn{2}{c}{ $ \text{conv}_{c}^{1 \times 1} + \text{Bias} $ } \\[1pt]
		\hline
		\multicolumn{2}{c}{ $ \text{Softmax} $ } \\
		\Xhline{3\arrayrulewidth}
	\end{tabularx} & \begin{tabularx}{0.48\linewidth}{!{\color{pool_red}\vrule width 2pt}Y!{\color{pool_red}\vrule width 2pt}!{\color{res_blue}\vrule width 2pt}Y!{\color{res_blue}\vrule width 2pt}}
		\Xhline{3\arrayrulewidth}
		\multicolumn{2}{c}{ FRRN B} \\					
		\hline
		\multicolumn{2}{c}{}\\[-8pt]
		\multicolumn{2}{c}{ $ \text{conv}_{48}^{5 \times 5} + \text{BN} + \text{ReLU} $ } \\[1pt]
		\hline
		\multicolumn{2}{c}{ $ 3 \times \text{RU}_{48}$ } \\
		\cellcolor{pool_red}\textcolor{white}{\footnotesize \textbf{pooling stream}} & \cellcolor{res_blue}\textcolor{white}{\footnotesize \textbf{residual stream}} \\
		& \\[-8pt]
		max pool & $\text{conv}_{32}^{1 \times 1}$ \\[1pt]
		\hline
		\multicolumn{2}{!{\color{pool_red}\vrule width 2pt}c!{\color{res_blue}\vrule width 2pt}}{ $3 \times \text{FRRU}_{96}$ } \\
		\hline
		max pool &  \\
		\hline
		\multicolumn{2}{!{\color{pool_red}\vrule width 2pt}c!{\color{res_blue}\vrule width 2pt}}{ $4 \times \text{FRRU}_{192}$ } \\
		\hline
		max pool &  \\
		\hline
		\multicolumn{2}{!{\color{pool_red}\vrule width 2pt}c!{\color{res_blue}\vrule width 2pt}}{ $2 \times \text{FRRU}_{384}$ } \\
		\hline
		max pool &  \\
		\hline
		\multicolumn{2}{!{\color{pool_red}\vrule width 2pt}c!{\color{res_blue}\vrule width 2pt}}{ $2 \times \text{FRRU}_{384}$ } \\
		\hline
		max pool &  \\
		\hline
		\multicolumn{2}{!{\color{pool_red}\vrule width 2pt}c!{\color{res_blue}\vrule width 2pt}}{ $2 \times \text{FRRU}_{384}$ } \\
		\hline
		unpool &  \\
		\hline
		\multicolumn{2}{!{\color{pool_red}\vrule width 2pt}c!{\color{res_blue}\vrule width 2pt}}{ $2 \times \text{FRRU}_{192}$ } \\
		\hline
		unpool &  \\
		\hline
		\multicolumn{2}{!{\color{pool_red}\vrule width 2pt}c!{\color{res_blue}\vrule width 2pt}}{ $2 \times \text{FRRU}_{192}$ } \\
		\hline
		unpool &  \\
		\hline
		\multicolumn{2}{!{\color{pool_red}\vrule width 2pt}c!{\color{res_blue}\vrule width 2pt}}{ $2 \times \text{FRRU}_{192}$ } \\
		\hline
		unpool &  \\
		\hline
		\multicolumn{2}{!{\color{pool_red}\vrule width 2pt}c!{\color{res_blue}\vrule width 2pt}}{ $2 \times \text{FRRU}_{96}$ } \\
		\hline
		unpool &  \\
		\cellcolor{pool_red}\textcolor{white}{\footnotesize \textbf{pooling stream}} & \cellcolor{res_blue}\textcolor{white}{\footnotesize \textbf{residual stream}} \\
		\multicolumn{2}{c}{ concatenate } \\
		\hline
		\multicolumn{2}{c}{ $3 \times \text{RU}_{48}$ } \\
		\hline
		\multicolumn{2}{c}{}\\[-8pt]
		\multicolumn{2}{c}{ $ \text{conv}_{c}^{1 \times 1} + \text{Bias} $ } \\[1pt]
		\hline
		\multicolumn{2}{c}{ $ \text{Softmax} $ } \\
		\Xhline{3\arrayrulewidth}
	\end{tabularx}
	\end{tabular}
	\label{tbl:networks}
\end{table}

Table \ref*{tbl:networks} shows the two network architectures that we used in order to assess our approach's segmentation performance.
The proposed architectures are based on several principles employed by other authors.
We follow Noh \etal \cite{Noh15ICCV} and use an encoder/decoder formulation.
In the encoder, we reduce the size of the pooling stream using max pooling operations.
The pooled feature maps are then successively upscaled using bilinear interpolation in the decoder.
Furthermore, similar to Simonyan and Zisserman \cite{Simonyan15ICLR}, we define a number of base channels that we double after each pooling operation (up to a certain upper limit).
Instead of choosing 64 base channels as in VGG net, we use 48 channels in order to have a manageable number of trainable parameters. Depending on the input image resolution, we use FRRN A or FRRN B to keep the relative size of the receptive fields consistent.

\section{Training Procedure}

\newcommand{\trainimline}[1]{\includegraphics[width=1\linewidth]{images/training/\imsize img_#1_16384} &
							\includegraphics[width=1\linewidth]{images/training/\imsize gt_#1_16384} &
							\includegraphics[width=1\linewidth]{images/training/\imsize prediction_#1_0}  &
							\includegraphics[width=1\linewidth]{images/training/\imsize prediction_#1_16384_white}  &
							\includegraphics[width=1\linewidth]{images/training/\imsize prediction_#1_32768_white}  &
							\includegraphics[width=1\linewidth]{images/training/\imsize prediction_#1_65536_white}  \\}

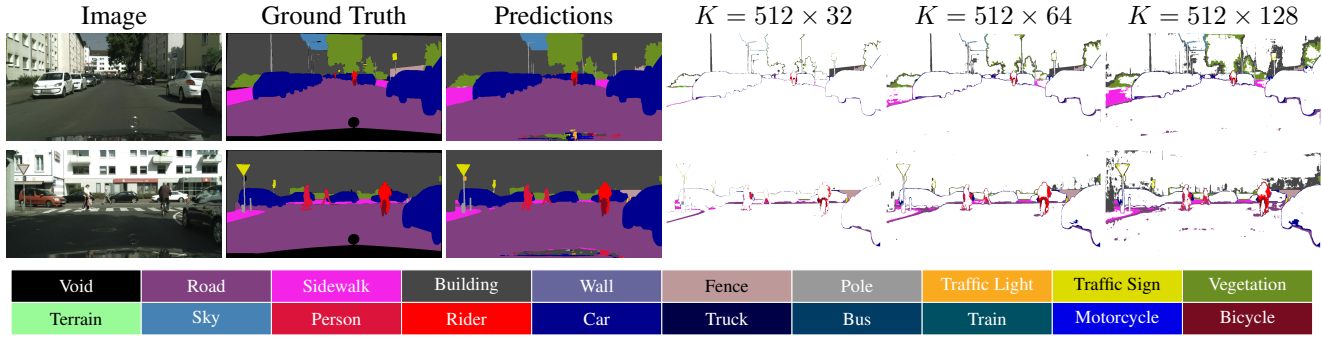
\begin{figure*}
\setlength\tabcolsep{1pt}
\begin{tabularx}{\linewidth}{@{}YYYYYY@{}}

Image & Ground Truth & Predictions & $K = 512 \times 32$ & $K = 512 \times 64$ & $K = 512 \times 128$  \\
%\trainimline{0}
\trainimline{1}
\trainimline{2}
\end{tabularx}
\resizebox{\linewidth}{!}{
\input{color_table2}
}
\caption{Pixels used by the bootstrapped cross-entropy loss for varying values of $K$.
The images and ground truth annotations originate from the twice-subsampled Cityscapes validation set \cite{Cordts16CVPR}.
Pixels that are labeled \emph{void} are not considered for the bootstrapping process.}
\label{fig:bootstrap}
\end{figure*}

Following Wu \etal, we train our network by minimizing a bootstrapped cross-entropy loss \cite{Wu16ARXIV}.
Let $c$ be the number of classes, $y_1,...,y_N \in \{1,...,c\}$ be the target class labels for the pixels $1,...,N$,
and let $p_{i,j}$ be the posterior class probability for class $j$ and pixel $i$. Then, the bootstrapped cross-entropy loss over $K$ pixels is defined as
\begin{align}
l = - \frac{1}{K} \sum_{i = 1}^N \mathbf{1}[p_{i, y_i} < t_K] \log p_{i, y_i} ~,
\end{align}
where $\mathbf{1}[x] = 1$ iff $x$ is true and $t_k \in \mathbb{R}$ is chosen such that $\left|\{i \in \{1,...,N\} : p_{i, y_i} < t_k \} \right| = K$.
The threshold parameter $t_k$ can easily be determined by sorting the predicted log probabilities and choosing the $K+1$-th one as threshold.
Figure \ref{fig:bootstrap} visualizes the concept.
Depending on the number of pixels $K$ that we consider, we select misclassified pixels or pixels where we predict the correct label with a small probability.
We minimize the loss using ADAM \cite{Kingma15ICLR}.

Because each FRRU processes features at the full image resolution, training a full-resolution residual network is very memory intensive.
Recall that in order for the backpropagation algorithm \cite{Rumelhart86Nature} to work, the entire forward pass has to be stored in memory.
If the memory required to store the forward pass for a given network exceeds the available GPU memory, we can no longer use the standard backpropagation algorithm.
In order to alleviate this problem, we partition the computation graph into several subsequent blocks by manually placing cut points in the graph.
We then compute the derivatives for each block individually.
To this end, we perform one (partial) forward pass per block and only store the feature maps for the block whose derivatives are computed given the derivative of the subsequent block.
This simple scheme allows us to manually control a space-time trade-off.
The idea of recomputing some intermediate results on demand is also used in \cite{Gruslys16ARXIV} and \cite{TChen16ARXIV}.
Note that these memory limitations only apply during training.
During testing, there is no need to store results of each operation in the network and our architecture's memory footprint is comparable to that of a ResNet encoder/decoder architecture.
We will make code for the gradient computation for arbitrary networks publicly available in Theano/Lasagne.

In order to reduce overfitting, we used two methods of data augmentation: \emph{translation augmentation} and \emph{gamma augmentation}.
The former method randomly translates an image and its annotations.
In order to keep consistent image dimensions, we have to pad the translated images and annotations.
To this end, we use reflection padding on the image and constant padding with void labels on the annotations.
Our second method of data augmentation is gamma augmentation. We use a slightly modified gamma augmentation method detailed in Appendix \ref{sec:appendix:gamma}.

\newcommand{\tabhead}[2]{\rotatebox{90}{\hspace*{-6pt} \mycirc[#1] \hspace*{-6pt} #2 \hspace*{2pt}}}
		\newcommand{\iouhead}[1]{\rotatebox{90}{\hspace*{-6pt} \ioucirc{} \hspace*{-6pt} #1 \hspace*{2pt}}}
		\newcommand{\rn}[1]{\num[round-mode=places,round-precision=1]{#1}}
		\newcommand{\rnb}[2]{\ifthenelse{\equal{#2}{1}}{\bfseries #1}{\ifthenelse{\equal{#2}{2}}{\bfseries\itshape#1}{#1}}}
		\newcolumntype{C}[1]{>{\centering\let\newline\\\arraybackslash\hspace{0pt}}m{#1}}
		\newcommand{\addtabline}[3]{\csvreader[	column count=48,
												%no head,
												late after line=\\,
												filter=\equal{\description}{#1}\and\equal{\coarse}{#2}
											]{data/external_baselines.csv}{6=\description, 3=\sample, 4=\coarse, 5=\nopretrain,
													7=\iou, 8=\road, 9=\sidewalk, 10=\building, 11=\wall, 12=\fence, 13=\pole, 14=\trafficlight, 15=\trafficsign, 16=\vegetation, 17=\terrain, 18=\sky, 19=\person, 20=\rider, 21=\car, 22=\truck, 23=\bus, 24=\train, 25=\motorcycle, 26=\bicycle, 28=\ioub, 29=\roadb,  30=\sidewalkb,  31=\buildingb, 32=\wallb, 33=\fenceb, 34=\poleb, 35=\trafficlightb, 36=\trafficsignb, 37=\vegetationb, 38=\terrainb, 39=\skyb, 40=\personb, 41=\riderb, 42=\carb, 43=\truckb, 44=\busb, 45=\trainb, 46=\motorcycleb, 47=\bicycleb}
											{#3 & {\tiny$\times$}\sample & \ifthenelse{\equal{\coarse}{1}}{\checkmark}{} & \ifthenelse{\equal{\nopretrain}{1}}{}{\checkmark} & \rnb{\iou}{\ioub} & \rnb{\road}{\roadb} & \rnb{\sidewalk}{\sidewalkb} & \rnb{\building}{\buildingb} & \rnb{\wall}{\wallb} & \rnb{\fence}{\fenceb} & \rnb{\pole}{\poleb} & \rnb{\trafficlight}{\trafficlightb} & \rnb{\trafficsign}{\trafficsignb} & \rnb{\vegetation}{\vegetationb} & \rnb{\terrain}{\terrainb} & \rnb{\sky}{\skyb} & \rnb{\person}{\personb} & \rnb{\rider}{\riderb} & \rnb{\car}{\carb} & \rnb{\truck}{\truckb} & \rnb{\bus}{\busb} & \rnb{\train}{\trainb} & \rnb{\motorcycle}{\motorcycleb} & \rnb{\bicycle}{\bicycleb}}}
		\newcommand{\addtablinegray}[3]{\csvreader[	column count=48,
												%no head,
												late after line=\\,
												filter=\equal{\description}{#1}\and\equal{\coarse}{#2}
											]{data/external_baselines.csv}{6=\description, 3=\sample, 4=\coarse, 5=\nopretrain,
													7=\iou, 8=\road, 9=\sidewalk, 10=\building, 11=\wall, 12=\fence, 13=\pole, 14=\trafficlight, 15=\trafficsign, 16=\vegetation, 17=\terrain, 18=\sky, 19=\person, 20=\rider, 21=\car, 22=\truck, 23=\bus, 24=\train, 25=\motorcycle, 26=\bicycle, 28=\ioub, 29=\roadb,  30=\sidewalkb,  31=\buildingb, 32=\wallb, 33=\fenceb, 34=\poleb, 35=\trafficlightb, 36=\trafficsignb, 37=\vegetationb, 38=\terrainb, 39=\skyb, 40=\personb, 41=\riderb, 42=\carb, 43=\truckb, 44=\busb, 45=\trainb, 46=\motorcycleb, 47=\bicycleb}
											{\cellcolor{gray_col}#3 & \cellcolor{gray_col}{\tiny$\times$}\sample &\cellcolor{gray_col} \ifthenelse{\equal{\coarse}{1}}{\checkmark}{} &\cellcolor{gray_col} \ifthenelse{\equal{\nopretrain}{1}}{}{\checkmark} &\cellcolor{gray_col} \rnb{\iou}{\ioub} & \cellcolor{gray_col}\rnb{\road}{\roadb} & \cellcolor{gray_col}\rnb{\sidewalk}{\sidewalkb} & \cellcolor{gray_col}\rnb{\building}{\buildingb} &\cellcolor{gray_col} \rnb{\wall}{\wallb} & \cellcolor{gray_col}\rnb{\fence}{\fenceb} &\cellcolor{gray_col} \rnb{\pole}{\poleb} &\cellcolor{gray_col} \rnb{\trafficlight}{\trafficlightb} &\cellcolor{gray_col} \rnb{\trafficsign}{\trafficsignb} &\cellcolor{gray_col} \rnb{\vegetation}{\vegetationb} & \cellcolor{gray_col}\rnb{\terrain}{\terrainb} & \cellcolor{gray_col}\rnb{\sky}{\skyb} &\cellcolor{gray_col} \rnb{\person}{\personb} &\cellcolor{gray_col} \rnb{\rider}{\riderb} & \cellcolor{gray_col}\rnb{\car}{\carb} & \cellcolor{gray_col}\rnb{\truck}{\truckb} & \cellcolor{gray_col}\rnb{\bus}{\busb} &\cellcolor{gray_col} \rnb{\train}{\trainb} &\cellcolor{gray_col} \rnb{\motorcycle}{\motorcycleb} &\cellcolor{gray_col} \rnb{\bicycle}{\bicycleb}}}

		\begin{table*}[t]
			\caption{	IoU scores from the cityscapes test set.
						We highlight the best published baselines for the different sampling rates.
						(Additional anonymous submissions exist as concurrent work.)
						Bold numbers represent the best, italic numbers the second best score for a class.
						We also indicate the subsampling factor used on the input images, whether additional coarsely annotated data was used, and whether the model was initialized with pre-trained weights. }
			\label{tab:result_test}
			\centering
			\small
			\setlength\tabcolsep{1.35pt}
			\begin{tabularx}{\textwidth}{Xp{1.3em}cp{1.3em}>{\centering}p{3em}cccccccccccccccccccc}
				\Xhline{3\arrayrulewidth}
				Method & \tabhead{white}{Subsample} & \tabhead{white}{Coarse} & \tabhead{white}{Pretrained} & \iouhead{Mean} & \tabhead{csroad}{Road} & \tabhead{cssidewalk}{Sidewalk} & \tabhead{csbuilding}{Building} & \tabhead{cswall}{Wall} & \tabhead{csfence}{Fence} & \tabhead{cspole}{Pole} & \tabhead{cstrafficlight}{Traf. Light} & \tabhead{cstrafficsign}{Traf. Sign} & \tabhead{csvegetation}{Vegetation} & \tabhead{csterrain}{Terrain} & \tabhead{cssky}{Sky} & \tabhead{csperson}{Person} & \tabhead{csrider}{Rider} & \tabhead{cscar}{Car} & \tabhead{cstruck}{Truck} & \tabhead{csbus}{Bus} & \tabhead{cstrain}{Train} & \tabhead{csmotorcycle}{Motorcycle} & \tabhead{csbicycle}{Bicycle}\\
				\Xhline{1.5\arrayrulewidth}
				\addtabline{"Segnet basic"}{0}{SegNet \cite{Badrinarayanan15aARXIV}}
				\addtabline{FRRN_small}{0}{FRRN A}
				\hline
				\addtabline{ENet}{0}{ENet \cite{Paszke16ARXIV}}
				\addtabline{"DeepLab LargeFOV StrongWeak"}{1}{DeepLab \cite{Papandreou15ICCV}}
				\addtablinegray{FRRN}{0}{FRRN B}
				\hline
				\addtabline{Dilation10}{0}{Dilation \cite{Yu16ICLR}}
				\addtabline{Adelaide_context}{0}{Adelaide \cite{Lin16CVPR}}
				\addtabline{LRR-4x}{0}{LRR \cite{Ghiasi16ECCV}}
				\addtabline{LRR-4x}{1}{LRR \cite{Ghiasi16ECCV}}
				\Xhline{3\arrayrulewidth}
			\end{tabularx}
		\end{table*}
\section{Experimental Evaluation}
We evaluate our approach on the recently released Cityscapes benchmark~\cite{Cordts16CVPR} containing images recorded in 50 different cities.
This benchmark provides 5,000 images with high-quality annotations split up into a training, validation, and test set (2,975, 500, and 1,525 images, respectively).
The dense pixel annotations span 30 classes frequently occurring in urban street scenes, out of which 19 are used for actual training and evaluation.
Annotations for the test set remain private and comparison to other methods is performed via a dedicated evaluation server.

We report the results of our FRRNs for two settings: FRRN A trained on quarter-resolution ($256 \times 512$) Cityscapes images; and FRRN B trained on half-resolution ($512 \times 1024$) images. 
We then upsample our predictions using bilinear interpolation in order to report scores at the full image resolution of $1024 \times 2048$ pixels.
Directly training at the full Cityscapes resolution turned out to be too memory intensive with our current design. However, as our experimental results will show, even when trained only on half-resolution images, our FRRN B's results are competitive with the best published methods trained on full-resolution data.
Unless specified otherwise, the reported results are based on the Cityscapes test set. Qualitative results are shown in Figure~\ref{fig:qualitative}, in Appendix \ref{sec:appendix:qualitativeresults}, and in our result video \footnote{https://www.youtube.com/watch?v=PNzQ4PNZSzc}.

\subsection{Residual Network Baseline}
Our network architecture can be described as a ResNet~\cite{He16CVPR} encoder/decoder architecture, where the residuals remain at the full input resolution throughout the network.
A natural baseline is thus a traditional ResNet encoder/decoder architecture with long-range skip connections~\cite{Long15CVPR,Noh15ICCV}.
In fact, such an architecture resembles a single deep hourglass module in the stacked hourglass network architecture~\cite{Newell16ECCV}.
This baseline differs from our proposed architecture in two important ways:
While the feature maps on our residual stream are processed by each FRRU, the feature maps on the long-range skip connections are not processed by intermediate layers.
Furthermore, long-range skip connections are scale dependent, meaning that features at one scale travel over a different skip connection than features at another scale.
This is in contrast to our network design, where the residual stream can carry upscaled features from several pooling stages simultaneously.

In order to illustrate the benefits of our approach over the natural baseline, we converted the architecture FRRN A (Table \ref{tbl:networks}a) to a ResNet as follows:
We first replaced all \mbox{FRRUs} by RUs and then added skip connections that connect the input of each pooling layer to the output of the corresponding unpooling layer.
The resulting ResNet has slightly fewer parameters than the original FRRN ($16.7\times 10^6$ vs. $17.7\times 10^6$).
This is due to the fact that RUs lack the $1\times1$ convolutions that connect the pooling to the residual stream.

We train both networks on the quarter-resolution Cityscapes dataset for 45,000 iterations at a batch size of 3.
We use a learning rate of $10^{-3}$ for the first 35,000 iterations and then reduce it to $10^{-4}$ for the following 10,000 iterations. 
Both networks converged within these iterations. 
The FRRN A resulted in a validation set mean IoU score of $65.7\%$ while the ResNet baseline only achieved $62.8\%$, showing a significant advantage of our FRRNs. 
Training FRRN B is performed in a similar fashion.
Detailed training curves are shown in Appendix \ref{sec:appendix:baseline}. 
% Figure \ref{X} show the mean IoU score on the validation set over time. 
% We can see that our model consistently outperforms the natural baseline with a significant margin. 

\subsection{Quantitative Evaluation}
\label{quantitative}

\PAR{Overview} In Table \ref{tab:result_test} we compare our method to the best (published) performers on the Cityscapes leader board, namely LRR~\cite{Ghiasi16ECCV}, Adelaide~\cite{Gu09CVPR}, and Dilation~\cite{Yu16ICLR}. 
Note that our network performs on par with the very complex and well engineered system by Ghiasi \etal (LRR). Among the top performers on Cityscapes, only ENet refrain from using a pre-trained network. However, they design their network for real time performance and thus do not obtain top scores. To the best of our knowledge, we are the first to show that it is possible to obtain state-of-the-art results even without pre-training. This gives credibility to our claim that network architectures can have a crucial effect on a system's overall performance.  

\PAR{Subsampling Factor.}
An interesting observation that we made on the Cityscapes test set is a correlation between the subsampling factor and the test performance.
This correlation can be seen in Figure~\ref{fig:iou_sampling} where we show the scores of several approaches currently listed on the leader board against their respective subsampling factors.
Unsurprisingly, most of the best performers operate on the full-resolution input images.
Throughout our experiments, we consistently outperformed other approaches who trained on the same image resolutions. 
Even though we only train on half-resolution images, Figure~\ref{fig:iou_sampling} clearly shows we can match the current published state-of-the-art (LRR~\cite{Ghiasi16ECCV}).
%(BL)
It is to be expected that further improvements can be obtained by switching to full-resolution training.

		\begin{figure}[t]
			\begin{tikzpicture}
				\begin{axis}[width=\linewidth,
									height=5cm,
									xlabel=Mean IoU Score (\%),
									ylabel=Subsampling factor,
									xmin=55, xmax=80,
									grid=both,
									major grid style={line width=.1pt, draw=gray!20},
									legend style={font=\tiny, at={(0.95, 0.95)},anchor=north east},
									ylabel style={font=\footnotesize},
									xlabel style={font=\footnotesize},
									xtick pos=left,
									ytick pos=left,
									ymin=0.75,
									ymax=4.25,
									xtick style={draw=none},
									every tick label/.append style={font=\footnotesize},
									scaled ticks=false,
									tick label style={/pgf/number format/fixed},
									legend columns=2,
									legend cell align={left},
									legend style={
										% the /tikz/ prefix is necessary here...
										% otherwise, it might end-up with `/pgfplots/column 2`
										% which is not what we want. compare pgfmanual.pdf
										/tikz/column 2/.style={
											column sep=5pt,
										},
									}]
					\addplot [black!50, only marks, discard if not={anon}{0}] table[x = iou, y = sample,col sep = comma]  {data/external_baselines.csv};
					\addplot [black!30, only marks, discard if not={anon}{1}] table[x = iou, y = sample,col sep = comma]  {data/external_baselines.csv};
					\addplot [lrr_col, only marks, discard if not and={bibkey}{Ghiasi16ECCV}{coarse}{1}] table[x = iou, y = sample,col sep = comma]  {data/external_baselines.csv};
					\addplot [adelaide_col, only marks, discard if not={bibkey}{Lin16CVPR}] table[x = iou, y = sample,col sep = comma]  {data/external_baselines.csv};
					\addplot [dilation_col, only marks, discard if not={bibkey}{Yu16ICLR}] table[x = iou, y = sample,col sep = comma]  {data/external_baselines.csv};
					\addplot [enet_col, only marks, discard if not={bibkey}{Paszke16ARXIV}] table[x = iou, y = sample,col sep = comma]  {data/external_baselines.csv};
					\addplot [segnet_col, only marks, discard if not={bibkey}{Badrinarayanan15aARXIV}] table[x = iou, y = sample,col sep = comma]  {data/external_baselines.csv};
					\addplot [deeplab_col, only marks, discard if not={bibkey}{Papandreou15ICCV}] table[x = iou, y = sample,col sep = comma]  {data/external_baselines.csv};
					\addplot [frrn_col, only marks, discard if not={description}{FRRN}] table[x = iou, y = sample,col sep = comma]  {data/external_baselines.csv};
					\addplot [frrn_col, only marks, discard if not={description}{FRRN_small}] table[x = iou, y = sample,col sep = comma]  {data/external_baselines.csv};
					\legend{Published, Unpublished, LRR \cite{Ghiasi16ECCV}, Adelaide \cite{Lin16CVPR}, Dilation \cite{Yu16ICLR}, ENet \cite{Paszke16ARXIV},SegNet \cite{Badrinarayanan15aARXIV}, DeepLab \cite{Papandreou15ICCV}, FRRN A/B}
				\end{axis}
			\end{tikzpicture}
			\caption{Comparison of the mean IoU scores of all approaches on the leader board of the Cityscapes segmentation benchmark based on the subsampling factor of the images that they were trained on.}
			\label{fig:iou_sampling}
		\end{figure}
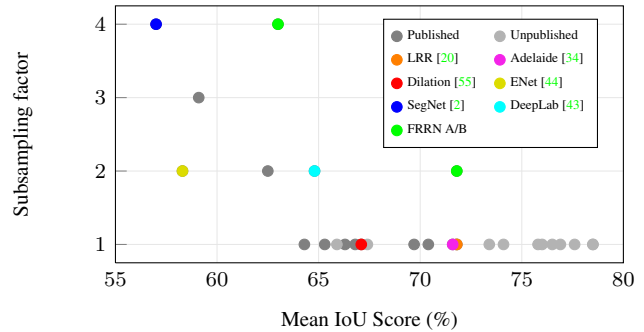

		\begin{figure}[t]
			\newcommand{\addtrimapplot}[3]{\addplot [discard if={tri_width}{-1}, #2] table[x = tri_width, y = #3,col sep = comma] {data/trimap/#1};}
			\begin{tikzpicture}
				\begin{axis}[width=\linewidth,
									height=5cm,
									xlabel={Trimap width $r$ [pixels]},
									ylabel={Mean IoU Score [\%]},
									xmin=0, xmax=80,
									grid=both,
									major grid style={line width=.1pt, draw=gray!20},
									legend style={font=\tiny, at={(0.95, 0.05)},anchor=south east},
									ylabel style={font=\footnotesize},
									xlabel style={font=\footnotesize},
									xtick pos=left,
									ytick pos=left,
									ymin=30,
									ymax=95,
									xtick style={draw=none},
									every tick label/.append style={font=\footnotesize},
									scaled ticks=false,
									tick label style={/pgf/number format/fixed},
									legend columns=2,
									legend cell align={left},
									legend style={
										% the /tikz/ prefix is necessary here...
										% otherwise, it might end-up with `/pgfplots/column 2`
										% which is not what we want. compare pgfmanual.pdf
										/tikz/column 2/.style={
											column sep=5pt,
										},
									}]
					\addtrimapplot{dilation.csv}{dilation_col}{iou}
					\addtrimapplot{lrr.csv}{lrr_col}{iou}
					\addtrimapplot{frrnd_halfres_no_coarse.csv}{frrn_col}{iou}
					\addtrimapplot{dilation_cat.csv}{dilation_col, dashed}{iou}
					\addtrimapplot{lrr_cat.csv}{lrr_col, dashed}{iou}
					\addtrimapplot{frrnd_halfres_no_coarse_cat.csv}{frrn_col, dashed}{iou}
					%\addtrimapplot{frrnd_halfres_no_coarse_cat.csv}{cscar}{vehicleiou}
					%\addtrimapplot{lrr_cat.csv}{cscar, dashed}{vehicleiou}
					%\addtrimapplot{dilation_cat.csv}{cscar, dotted}{vehicleiou}
					%\addtrimapplot{frrnd_halfres_no_coarse_cat.csv}{csperson}{humaniou}
					%\addtrimapplot{lrr_cat.csv}{csperson, dashed}{humaniou}
					%\addtrimapplot{dilation_cat.csv}{csperson, dotted}{humaniou}
					%\addtrimapplot{frrnd_halfres_no_coarse_cat.csv}{csbuilding}{constructioniou}
					%\addtrimapplot{lrr_cat.csv}{csbuilding, dashed}{constructioniou}
					%\addtrimapplot{dilation_cat.csv}{csbuilding, dotted}{constructioniou}
					%\addtrimapplot{frrnd_halfres_no_coarse_cat.csv}{csroad}{flatiou}
					%\addtrimapplot{lrr_cat.csv}{csroad, dashed}{flatiou}
					%\addtrimapplot{dilation_cat.csv}{csroad, dotted}{flatiou}
					%\legend{FRRN class, FRRN cat, LRR class \cite{Ghiasi16ECCV}, LRR cat \cite{Ghiasi16ECCV}, Dilation class \cite{Yu16ICLR}, Dilation cat \cite{Yu16ICLR}}
					\legend{Dilation \cite{Yu16ICLR}, LRR \cite{Ghiasi16ECCV},FRRN B}
				\end{axis}
			\end{tikzpicture}
			\caption{
				The trimap evaluation on the validation set.
				The solid lines show the mean IoU score of our approach and two top performing methods that released their code.
				The dashed lines show the mean IoU score when using the 7 Cityscapes category labels for the same methods.
			}
			\label{fig:trimap}
		\end{figure}
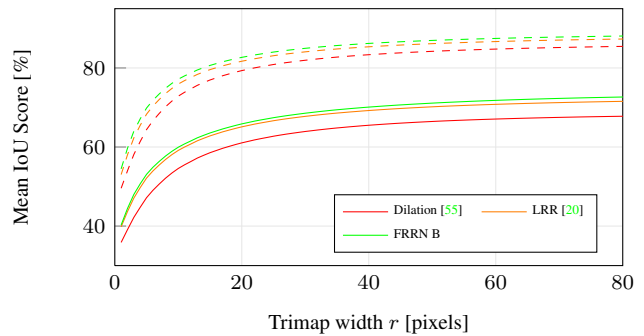

		\newcommand{\resimline}[1]{\includegraphics[width=1\linewidth]{images/results/ims/\imsize#1}&
							\includegraphics[width=1\linewidth]{images/results/gt/\imsize#1}&
							\includegraphics[width=1\linewidth]{images/results/ours/\imsize#1}&
							\includegraphics[width=1\linewidth]{images/results/lrr/\imsize#1}&
							\includegraphics[width=1\linewidth]{images/results/dilation/\imsize#1}\\}

		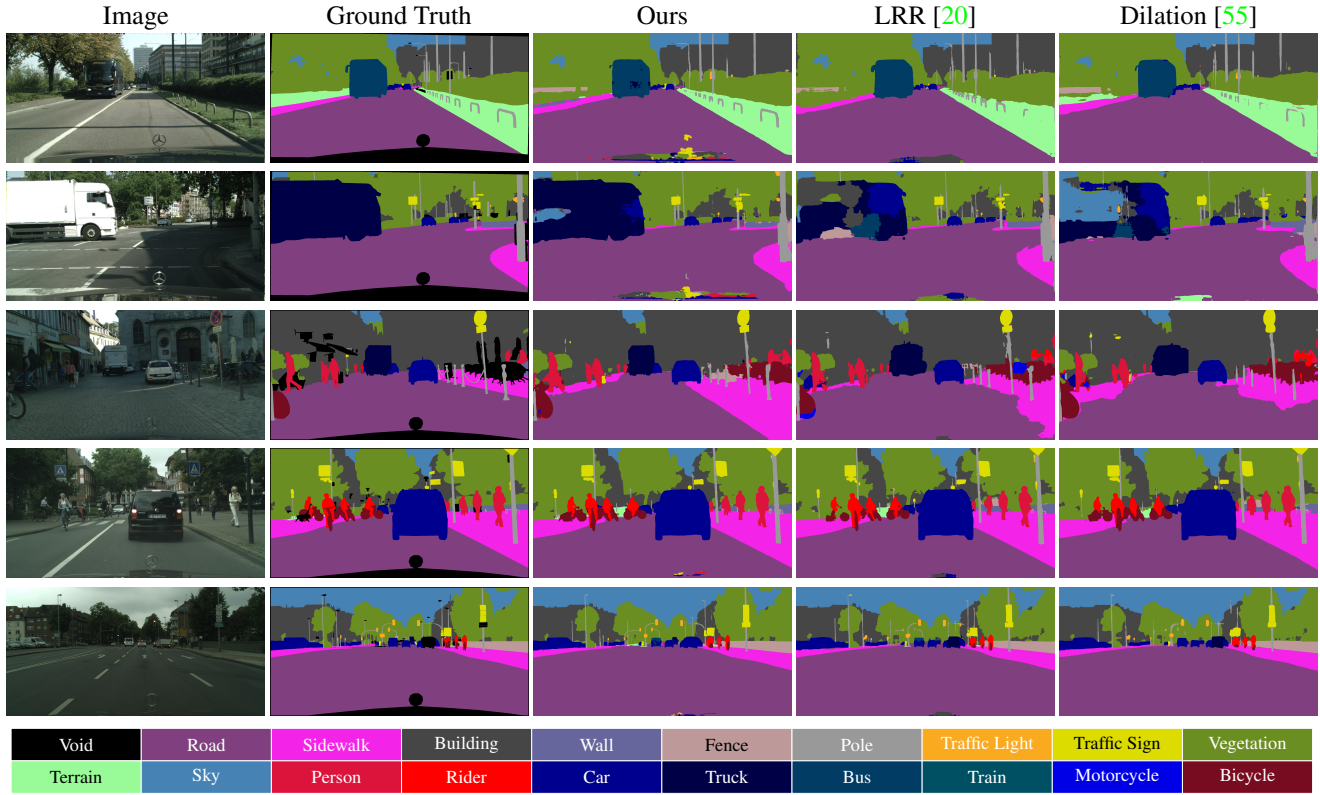
\begin{figure*}
		\setlength\tabcolsep{1pt}
		\begin{tabularx}{\linewidth}{@{}YYYYYY@{}}
		Image & Ground Truth & Ours & LRR \cite{Ghiasi16ECCV} & Dilation \cite{Yu16ICLR}\\
		\resimline{frankfurt_000000_007365}
		\resimline{frankfurt_000000_010351}
		%\resimline{frankfurt_000001_077233}
		%\resimline{lindau_000018_000019}
		\resimline{lindau_000024_000019}
		%\resimline{munster_000065_000019}
		\resimline{munster_000055_000019}
		\resimline{munster_000129_000019}
		%\resimline{munster_000097_000019}
		%\resimline{munster_000128_000019}
		%\resimline{munster_000157_000019}
		%\resimline{munster_000167_000019}
		\end{tabularx}
		\resizebox{\linewidth}{!}{
			\input{color_table2}
		}
		\caption{
			Qualitative comparison on the Cityscapes validation set. Interesting cases are the fence in the first row, the truck in the second row, or the street light poles in the last row. An interesting failure case is shown in the third row: all methods struggle to find the correct sidewalk boundary, however our network makes a clean and reasonable prediction. 
		}
		\label{fig:qualitative}
		\end{figure*}

\subsection{Boundary Adherence}
\vspace*{-3pt}
Due to several pooling operations (and subsequent upsampling) in many of today's FCN architectures, boundaries are often overly smooth, resulting in lost details and edge-bleeding.
This leads to suboptimal scores, but it also makes the output of a semantic segmentation approach harder to use without further post-processing.
Since inaccurate boundaries are often not apparent from the standard evaluation metric scores, a typical approach is a trimap evaluation in order to quantify detailed boundary adherence~\cite{Kohli09IJCV,Krahenbuhl11NIPS,Ghiasi16ECCV}.
During trimap evaluation, all predictions are ignored if they do not fall within a certain radius $r$ of a ground truth label boundary.
Figure~\ref{fig:trimap} visualizes our trimap evaluation performed on the validation set for varying trimap widths $r$ between 1 and 80 pixels.
We compare to LRR~\cite{Ghiasi16ECCV} and Dilation~\cite{Yu16ICLR}, who made code and pre-trained models available.
We see that our approach outperforms the competition consistently for all radii $r$. 
Furthermore, it shall be noted that the method of~\cite{Ghiasi16ECCV} is based on an architecture specifically designed for clean boundaries.
Our method achieves better boundary adherence, both numerically and qualitatively (see Figure~\ref{fig:qualitative}), with a much simpler architecture and without ImageNet pre-training.

Often one can boost both the numerical score and the boundary adherence by using a fully connected CRF as post-processing step.
We tried to apply a fully connected CRF with Gaussian kernel, as introduced by Kr{\"{a}}henb{\"{u}}hl and Kolton~\cite{Krahenbuhl11NIPS}.
We used the standard appearance and smoothness kernels and tuned parameters on the validation set by running several thousand Hyperopt iterations~\cite{Bergstra13ICML}.
Surprisingly the color standard deviation for the appearance kernel tended towards very small values, while the weight did not go to zero.
This indicates that the appearance kernel would only smooth labels across pixels with very similar colors.
Nevertheless, with the best parameters we only obtained an IoU boost of $\sim0.5\%$ on the validation set.
Given the high computation time we decided against any post-processing steps.

\section{Conclusion}
\vspace*{-3pt}
In this paper we propose a novel network architecture for semantic segmentation in street scenes. Our architecture is clean, does not require additional post-processing, can be trained from scratch, shows superior boundary adherence, and reaches state-of-the-art results on the Cityscapes benchmark. We will provide code and all trained models. Since we do not incorporate design choices specifically tailored towards semantic segmentation, we believe that our architecture will also be applicable to other tasks such as stereo or optical flow where predictions are performed per pixel.

{\small
\bibliographystyle{ieee}
\bibliography{abbrev_short,egbib}
}

\clearpage

\begin{appendix}

\noindent\begin{Large}
\hspace{-7.8pt}
\textbf{Appendix}
\end{Large}

\section{Gamma Augmentation}
\label{sec:appendix:gamma}

			\newcommand{\samplevar}{50}
			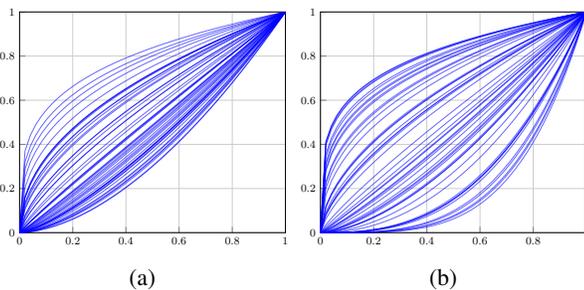
\begin{figure}[b]
				\begin{subfigure}{0.48\linewidth}
				\centering
					\resizebox{\linewidth}{!}{
					\begin{tikzpicture}[tight background]
						\begin{axis}[xmin=0, xmax=1, ymin=0, ymax=1, enlargelimits=false,
								grid=both,
								major grid style={line width=.1pt, draw=gray!40},
								ylabel style={font=\footnotesize},
								xlabel style={font=\footnotesize},
								xtick pos=left,
								ytick pos=left,
								every tick label/.append style={font=\footnotesize}, ]
						\addplot[blue, ultra thin, opacity=0.6, domain=0:1, samples=\samplevar] {pow(x, 1.128500)};
						\addplot[blue, ultra thin, opacity=0.6, domain=0:1, samples=\samplevar] {pow(x, 1.595996)};
						\addplot[blue, ultra thin, opacity=0.6, domain=0:1, samples=\samplevar] {pow(x, 0.515135)};
						\addplot[blue, ultra thin, opacity=0.6, domain=0:1, samples=\samplevar] {pow(x, 0.565716)};
						\addplot[blue, ultra thin, opacity=0.6, domain=0:1, samples=\samplevar] {pow(x, 1.102926)};
						\addplot[blue, ultra thin, opacity=0.6, domain=0:1, samples=\samplevar] {pow(x, 0.925790)};
						\addplot[blue, ultra thin, opacity=0.6, domain=0:1, samples=\samplevar] {pow(x, 0.939443)};
						\addplot[blue, ultra thin, opacity=0.6, domain=0:1, samples=\samplevar] {pow(x, 1.054726)};
						\addplot[blue, ultra thin, opacity=0.6, domain=0:1, samples=\samplevar] {pow(x, 0.255801)};
						\addplot[blue, ultra thin, opacity=0.6, domain=0:1, samples=\samplevar] {pow(x, 0.738766)};
						\addplot[blue, ultra thin, opacity=0.6, domain=0:1, samples=\samplevar] {pow(x, 1.448811)};
						\addplot[blue, ultra thin, opacity=0.6, domain=0:1, samples=\samplevar] {pow(x, 1.297795)};
						\addplot[blue, ultra thin, opacity=0.6, domain=0:1, samples=\samplevar] {pow(x, 0.477621)};
						\addplot[blue, ultra thin, opacity=0.6, domain=0:1, samples=\samplevar] {pow(x, 0.948280)};
						\addplot[blue, ultra thin, opacity=0.6, domain=0:1, samples=\samplevar] {pow(x, 0.356212)};
						\addplot[blue, ultra thin, opacity=0.6, domain=0:1, samples=\samplevar] {pow(x, 0.465451)};
						\addplot[blue, ultra thin, opacity=0.6, domain=0:1, samples=\samplevar] {pow(x, 0.537862)};
						\addplot[blue, ultra thin, opacity=0.6, domain=0:1, samples=\samplevar] {pow(x, 1.737717)};
						\addplot[blue, ultra thin, opacity=0.6, domain=0:1, samples=\samplevar] {pow(x, 1.521992)};
						\addplot[blue, ultra thin, opacity=0.6, domain=0:1, samples=\samplevar] {pow(x, 0.855151)};
						\addplot[blue, ultra thin, opacity=0.6, domain=0:1, samples=\samplevar] {pow(x, 0.815888)};
						\addplot[blue, ultra thin, opacity=0.6, domain=0:1, samples=\samplevar] {pow(x, 1.044113)};
						\addplot[blue, ultra thin, opacity=0.6, domain=0:1, samples=\samplevar] {pow(x, 0.577801)};
						\addplot[blue, ultra thin, opacity=0.6, domain=0:1, samples=\samplevar] {pow(x, 0.467020)};
						\addplot[blue, ultra thin, opacity=0.6, domain=0:1, samples=\samplevar] {pow(x, 0.656226)};
						\addplot[blue, ultra thin, opacity=0.6, domain=0:1, samples=\samplevar] {pow(x, 0.740813)};
						\addplot[blue, ultra thin, opacity=0.6, domain=0:1, samples=\samplevar] {pow(x, 0.653977)};
						\addplot[blue, ultra thin, opacity=0.6, domain=0:1, samples=\samplevar] {pow(x, 1.394314)};
						\addplot[blue, ultra thin, opacity=0.6, domain=0:1, samples=\samplevar] {pow(x, 0.470037)};
						\addplot[blue, ultra thin, opacity=0.6, domain=0:1, samples=\samplevar] {pow(x, 1.386796)};
						\addplot[blue, ultra thin, opacity=0.6, domain=0:1, samples=\samplevar] {pow(x, 0.408322)};
						\addplot[blue, ultra thin, opacity=0.6, domain=0:1, samples=\samplevar] {pow(x, 1.082375)};
						\addplot[blue, ultra thin, opacity=0.6, domain=0:1, samples=\samplevar] {pow(x, 1.033171)};
						\addplot[blue, ultra thin, opacity=0.6, domain=0:1, samples=\samplevar] {pow(x, 0.320815)};
						\addplot[blue, ultra thin, opacity=0.6, domain=0:1, samples=\samplevar] {pow(x, 1.202513)};
						\addplot[blue, ultra thin, opacity=0.6, domain=0:1, samples=\samplevar] {pow(x, 1.249883)};
						\addplot[blue, ultra thin, opacity=0.6, domain=0:1, samples=\samplevar] {pow(x, 1.169531)};
						\addplot[blue, ultra thin, opacity=0.6, domain=0:1, samples=\samplevar] {pow(x, 1.586669)};
						\addplot[blue, ultra thin, opacity=0.6, domain=0:1, samples=\samplevar] {pow(x, 0.292345)};
						\addplot[blue, ultra thin, opacity=0.6, domain=0:1, samples=\samplevar] {pow(x, 0.518001)};
						\addplot[blue, ultra thin, opacity=0.6, domain=0:1, samples=\samplevar] {pow(x, 1.343989)};
						\addplot[blue, ultra thin, opacity=0.6, domain=0:1, samples=\samplevar] {pow(x, 0.689026)};
						\addplot[blue, ultra thin, opacity=0.6, domain=0:1, samples=\samplevar] {pow(x, 0.925999)};
						\addplot[blue, ultra thin, opacity=0.6, domain=0:1, samples=\samplevar] {pow(x, 1.539841)};
						\addplot[blue, ultra thin, opacity=0.6, domain=0:1, samples=\samplevar] {pow(x, 1.630720)};
						\addplot[blue, ultra thin, opacity=0.6, domain=0:1, samples=\samplevar] {pow(x, 1.343801)};
						\addplot[blue, ultra thin, opacity=0.6, domain=0:1, samples=\samplevar] {pow(x, 1.724412)};
						\addplot[blue, ultra thin, opacity=0.6, domain=0:1, samples=\samplevar] {pow(x, 1.337520)};
						\addplot[blue, ultra thin, opacity=0.6, domain=0:1, samples=\samplevar] {pow(x, 1.225040)};
						\addplot[blue, ultra thin, opacity=0.6, domain=0:1, samples=\samplevar] {pow(x, 1.358421)};
						\end{axis}
					\end{tikzpicture}}
					\caption{}
					\label{fig:gamma_a}
				\end{subfigure}%
				\begin{subfigure}{0.48\linewidth}
				\centering
					\resizebox{\linewidth}{!}{
					\begin{tikzpicture}[tight background]
						\begin{axis}[xmin=0, xmax=1, ymin=0, ymax=1, enlargelimits=false,
								grid=both,
								major grid style={line width=.1pt, draw=gray!40},
								ylabel style={font=\footnotesize},
								xlabel style={font=\footnotesize},
								xtick pos=left,
								ytick pos=left,
								every tick label/.append style={font=\footnotesize}, ]
						\addplot[blue, ultra thin, opacity=0.6, domain=0:1, samples=\samplevar] {pow(x, 0.234318)};
						\addplot[blue, ultra thin, opacity=0.6, domain=0:1, samples=\samplevar] {pow(x, 0.412055)};
						\addplot[blue, ultra thin, opacity=0.6, domain=0:1, samples=\samplevar] {pow(x, 3.340263)};
						\addplot[blue, ultra thin, opacity=0.6, domain=0:1, samples=\samplevar] {pow(x, 1.632278)};
						\addplot[blue, ultra thin, opacity=0.6, domain=0:1, samples=\samplevar] {pow(x, 2.717150)};
						\addplot[blue, ultra thin, opacity=0.6, domain=0:1, samples=\samplevar] {pow(x, 1.499970)};
						\addplot[blue, ultra thin, opacity=0.6, domain=0:1, samples=\samplevar] {pow(x, 1.280350)};
						\addplot[blue, ultra thin, opacity=0.6, domain=0:1, samples=\samplevar] {pow(x, 1.064177)};
						\addplot[blue, ultra thin, opacity=0.6, domain=0:1, samples=\samplevar] {pow(x, 0.555242)};
						\addplot[blue, ultra thin, opacity=0.6, domain=0:1, samples=\samplevar] {pow(x, 1.161980)};
						\addplot[blue, ultra thin, opacity=0.6, domain=0:1, samples=\samplevar] {pow(x, 1.437325)};
						\addplot[blue, ultra thin, opacity=0.6, domain=0:1, samples=\samplevar] {pow(x, 0.551097)};
						\addplot[blue, ultra thin, opacity=0.6, domain=0:1, samples=\samplevar] {pow(x, 1.207614)};
						\addplot[blue, ultra thin, opacity=0.6, domain=0:1, samples=\samplevar] {pow(x, 2.741176)};
						\addplot[blue, ultra thin, opacity=0.6, domain=0:1, samples=\samplevar] {pow(x, 1.668427)};
						\addplot[blue, ultra thin, opacity=0.6, domain=0:1, samples=\samplevar] {pow(x, 0.302076)};
						\addplot[blue, ultra thin, opacity=0.6, domain=0:1, samples=\samplevar] {pow(x, 2.107278)};
						\addplot[blue, ultra thin, opacity=0.6, domain=0:1, samples=\samplevar] {pow(x, 1.177167)};
						\addplot[blue, ultra thin, opacity=0.6, domain=0:1, samples=\samplevar] {pow(x, 0.637523)};
						\addplot[blue, ultra thin, opacity=0.6, domain=0:1, samples=\samplevar] {pow(x, 0.375384)};
						\addplot[blue, ultra thin, opacity=0.6, domain=0:1, samples=\samplevar] {pow(x, 0.855881)};
						\addplot[blue, ultra thin, opacity=0.6, domain=0:1, samples=\samplevar] {pow(x, 3.791002)};
						\addplot[blue, ultra thin, opacity=0.6, domain=0:1, samples=\samplevar] {pow(x, 2.848499)};
						\addplot[blue, ultra thin, opacity=0.6, domain=0:1, samples=\samplevar] {pow(x, 0.384804)};
						\addplot[blue, ultra thin, opacity=0.6, domain=0:1, samples=\samplevar] {pow(x, 0.275975)};
						\addplot[blue, ultra thin, opacity=0.6, domain=0:1, samples=\samplevar] {pow(x, 0.607875)};
						\addplot[blue, ultra thin, opacity=0.6, domain=0:1, samples=\samplevar] {pow(x, 0.265034)};
						\addplot[blue, ultra thin, opacity=0.6, domain=0:1, samples=\samplevar] {pow(x, 0.638853)};
						\addplot[blue, ultra thin, opacity=0.6, domain=0:1, samples=\samplevar] {pow(x, 3.452812)};
						\addplot[blue, ultra thin, opacity=0.6, domain=0:1, samples=\samplevar] {pow(x, 0.319410)};
						\addplot[blue, ultra thin, opacity=0.6, domain=0:1, samples=\samplevar] {pow(x, 1.076541)};
						\addplot[blue, ultra thin, opacity=0.6, domain=0:1, samples=\samplevar] {pow(x, 0.295365)};
						\addplot[blue, ultra thin, opacity=0.6, domain=0:1, samples=\samplevar] {pow(x, 0.654122)};
						\addplot[blue, ultra thin, opacity=0.6, domain=0:1, samples=\samplevar] {pow(x, 3.693893)};
						\addplot[blue, ultra thin, opacity=0.6, domain=0:1, samples=\samplevar] {pow(x, 0.227724)};
						\addplot[blue, ultra thin, opacity=0.6, domain=0:1, samples=\samplevar] {pow(x, 1.388102)};
						\addplot[blue, ultra thin, opacity=0.6, domain=0:1, samples=\samplevar] {pow(x, 2.709128)};
						\addplot[blue, ultra thin, opacity=0.6, domain=0:1, samples=\samplevar] {pow(x, 1.664260)};
						\addplot[blue, ultra thin, opacity=0.6, domain=0:1, samples=\samplevar] {pow(x, 0.327375)};
						\addplot[blue, ultra thin, opacity=0.6, domain=0:1, samples=\samplevar] {pow(x, 0.454886)};
						\addplot[blue, ultra thin, opacity=0.6, domain=0:1, samples=\samplevar] {pow(x, 0.227395)};
						\addplot[blue, ultra thin, opacity=0.6, domain=0:1, samples=\samplevar] {pow(x, 1.013165)};
						\addplot[blue, ultra thin, opacity=0.6, domain=0:1, samples=\samplevar] {pow(x, 1.461112)};
						\addplot[blue, ultra thin, opacity=0.6, domain=0:1, samples=\samplevar] {pow(x, 0.936760)};
						\addplot[blue, ultra thin, opacity=0.6, domain=0:1, samples=\samplevar] {pow(x, 0.237725)};
						\addplot[blue, ultra thin, opacity=0.6, domain=0:1, samples=\samplevar] {pow(x, 2.636775)};
						\addplot[blue, ultra thin, opacity=0.6, domain=0:1, samples=\samplevar] {pow(x, 4.002555)};
						\addplot[blue, ultra thin, opacity=0.6, domain=0:1, samples=\samplevar] {pow(x, 1.869112)};
						\addplot[blue, ultra thin, opacity=0.6, domain=0:1, samples=\samplevar] {pow(x, 0.711069)};
						\addplot[blue, ultra thin, opacity=0.6, domain=0:1, samples=\samplevar] {pow(x, 0.541572)};
						\end{axis}
					\end{tikzpicture}}
					\caption{}
					\label{fig:gamma_b}
				\end{subfigure}%
				\caption{
					Both plots show the function $x \mapsto x^\gamma$ for 50 samples of $\gamma$.
					In plot (a), $\gamma$ is sampled uniformly at random from the interval $[0.25, 1.75]$.
					%In plot (b), we apply the nonlinear transformation $\gamma = \frac{\log(0.5 + 2^{-0.5} Z)}{\log(0.5 - 2^{-0.5} Z)}$ where $Z$ is sampled uniformly at random from the interval $[-a, a]$ for $a = 0.35$, which reduces the bias.
					In plot (b), we use Equation \ref{eq:63} where $Z$ is sampled uniformly from the interval $[-0.35, 0.35]$.
					Our new sampling reduces the bias.
					}
				\label{fig:gamma}
				\vspace{5pt}
			\end{figure}
			
Gamma augmentation is an augmentation method that varies the image contrast and brightness.
Assume the intensity values of an image are scaled to the unit interval $[0, 1]$.
Then gamma augmentation applies the intensity transformation $x \mapsto x^\gamma$ for a randomly sampled augmentation parameter $\gamma > 0$. 
However, sampling the augmentation parameter $\gamma$ is not trivial.
Naively drawing samples from a uniform or truncated Gaussian distribution with a mean of $1$ results in a noticeable bias (Figure \ref{fig:gamma_a}).
In order to reduce the bias, we deduce a novel sampling schema for $\gamma$. 

Let $U$ be a random variable that is implicitly defined as the solution to the fixed-point problem
\begin{align}
  \label{eq:61}
  1 - U^\gamma = U.
\end{align}
Our goal is to find $\gamma$ such that $\mathbb{E}_U[U] = 0.5$. 
The key idea to solving this problem is to look at the deviation of $U$ from $0.5$. 
Let $Z$ be this deviation. Then (\ref{eq:61}) is equivalent to 
\begin{align}
  \label{eq:62}
  \left(0.5 - \frac{1}{\sqrt{2}}Z \right)^\gamma = 0.5 + \frac{1}{\sqrt{2}}Z.
\end{align}
Now $\mathbb{E}_U[U] = 0.5$ implies $\mathbb{E}_Z[Z] = 0$ and solving (\ref{eq:62}) for $\gamma$ yields
\begin{align}
  \label{eq:63}
  \gamma = \frac{\log\left(0.5 + 2^{-0.5} Z\right)}{\log \left(0.5 - 2^{-0.5} Z\right)}.
\end{align}
Hence, without solving for the implicitly defined variable $U$ explicitly, we found a transformation of a zero-mean random variable $Z$ such that $\gamma$ has the desired properties. 
Because $Z$ was defined to be the offset from $0.5$ and $U \in [0, 1]$, it follows $Z \in [-0.5, 0.5]$.
We are free to choose any distribution such that $Z$ has zero mean and falls into the range $[-0.5, 0.5]$. 
For simplicity reasons, we choose $Z$ to be uniformly distributed over $[-a, a]$ where $a \in [0, 0.5]$ determines the strength of the augmentation. 
Figure \ref{fig:gamma_b} illustrates the obvious bias reduction.

%%%%%%%%% BODY TEXT
\section{Baseline Evaluation}
\label{sec:appendix:baseline}
\pgfplotstableread[col sep=tab]{baseline.csv}\baselinedata
\begin{figure}[t]
	\resizebox{\linewidth}{!}{
	\begin{tikzpicture}[tight background]
		\begin{axis}[
   width=\linewidth,
   xlabel=Iteration,
   ylabel=IoU score,
	       	grid=both, 
	       	major grid style={line width=.1pt, draw=gray!40}, 
	       	legend style={font=\tiny, at={(0.95, 0.05)},anchor=south east},
	       	legend columns=1,
	       	ylabel style={font=\footnotesize}, 
	       	xlabel style={font=\footnotesize}, 
	       	height=6cm, 
	       	xtick pos=left, 
	       	ytick pos=left, 
	       	xtick style={draw=none},  
	       	every tick label/.append style={font=\footnotesize}, 
	       	xmin=0, 
	       	ymin=0, 
	       	ymax=80, 
	       	xmax=45000,
	       	scaled ticks=false, 
	       	tick label style={/pgf/number format/fixed}
	       	]
	       	
			\addplot[blue, dashed, no marks, y filter/.code={\pgfmathparse{\pgfmathresult*100}\pgfmathresult}] table[x index=0, y index=1] \baselinedata;
			\addplot[red, dashed, no marks, y filter/.code={\pgfmathparse{\pgfmathresult*100}\pgfmathresult}] table[x index=0, y index=2] \baselinedata;
			\addplot[blue, no marks, y filter/.code={\pgfmathparse{\pgfmathresult*100}\pgfmathresult}] table[x index=0, y index=3] \baselinedata;
			\addplot[red, no marks, y filter/.code={\pgfmathparse{\pgfmathresult*100}\pgfmathresult}] table[x index=0, y index=4] \baselinedata;
			
			\legend{FRRN A, ResNet with long-range skips, current best FRRN A, current best ResNet}
		\end{axis}
	\end{tikzpicture}} 
	\caption{
		The plot shows the IoU score on the Cityscapes validation set as a function of the number of training iterations for the baseline architecture and FRRN A.
		The solid lines show the best IoU score up to iteration $N$.
	}
	\label{fig:baseline}
\end{figure}
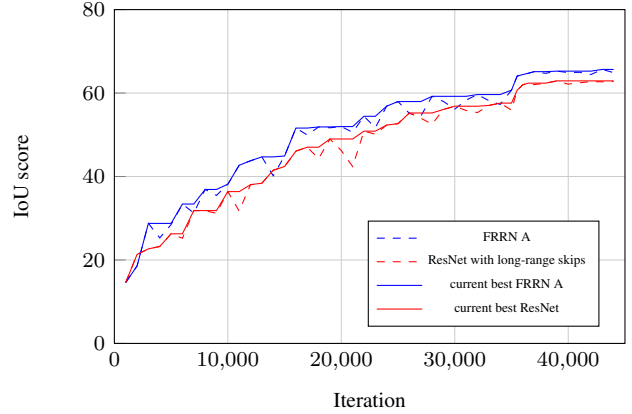
In Section 5.2 of the main paper, we describe the setting of our baseline method (Residual Network Baseline) and compare it to our FRRN A network. 
To emphasize on a proper training procedure of both baselines, Figure \ref{fig:baseline} shows the mean IoU score on the validation set over time. 
We can see that our model outperforms the baseline with a significant margin and both methods are trained until convergence.

\newcommand{\resimlinetwo}[1]{\includegraphics[width=1\linewidth]{images/results/ims/\imsize#1}&
          \includegraphics[width=1\linewidth]{images/results/gt/\imsize#1}&
          \includegraphics[width=1\linewidth]{images/results/ours/\imsize#1}&
          \includegraphics[width=1\linewidth]{images/results/lrr/\imsize#1}\\[-2pt]}

\begin{figure*}[t]
    \setlength\tabcolsep{1pt}
    \begin{tabularx}{\linewidth}{@{}YYYYYY@{}}
      Image & Ground Truth & Ours & LRR \cite{Ghiasi16ECCV}\\
      %\resimlinetwo{frankfurt_000000_007365}
      %\resimlinetwo{frankfurt_000000_010351}
      \resimlinetwo{frankfurt_000001_077233}
      \resimlinetwo{lindau_000018_000019}
      %\resimlinetwo{lindau_000024_000019}
      \resimlinetwo{munster_000065_000019}
      %\resimlinetwo{munster_000055_000019}
      %\resimlinetwo{munster_000129_000019}
      \resimlinetwo{munster_000097_000019}
      \resimlinetwo{munster_000128_000019}
      \resimlinetwo{munster_000157_000019}
      \resimlinetwo{munster_000167_000019}
      \resimlinetwo{frankfurt_000001_058914}
    \end{tabularx}
    \resizebox{\linewidth}{!}{
      \input{color_table2}
    }
    \caption{
      Additional qualitative results on the Cityscapes validation set.
      We omit the comparison to Dilation \cite{Yu16ICLR} in order to show bigger images here.
    }
    \label{fig:qualitative_appendix}
\end{figure*}
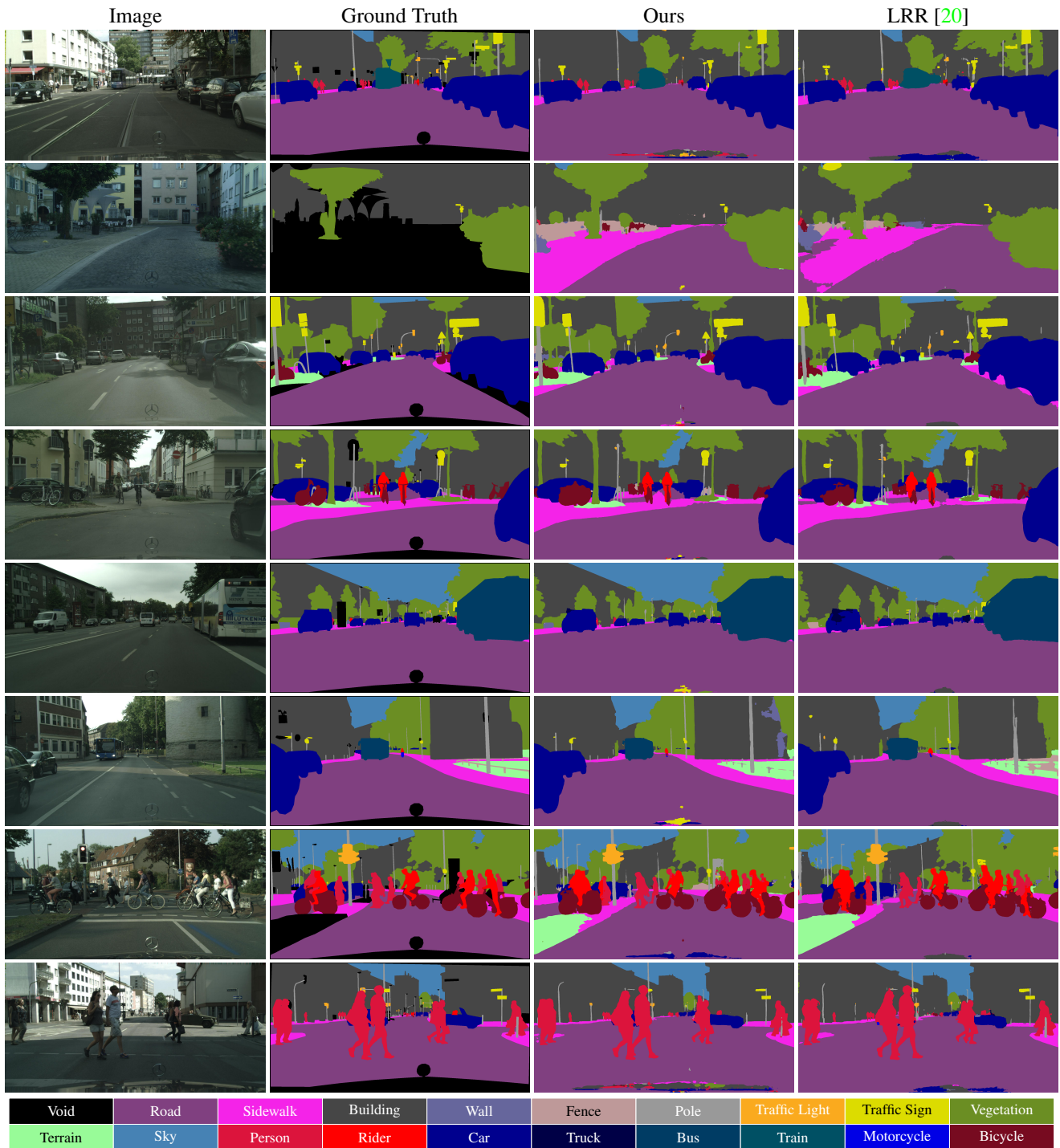

\section{Qualitative Results}
\label{sec:appendix:qualitativeresults}
Figure \ref{fig:qualitative_appendix} shows and compares addtional output labelings of our method.
Please also consult our labeled video sequence\footnote{https://www.youtube.com/watch?v=PNzQ4PNZSzc} to gain a better sense of the quality of our method.
\textcolor{white}{We all know Latex is a pain.}

\newpage
\end{appendix}
\end{document}

%% file: nn_styles.tex
\definecolor{nn-input-color}{RGB}{187,187,187}
\definecolor{nn-fc-color}{RGB}{251,255,123}
\definecolor{nn-conv-color}{RGB}{208,225,166}
\definecolor{nn-bias-color}{RGB}{225,222,166}
\definecolor{nn-bn-color}{RGB}{243,140,136}
\definecolor{nn-relu-color}{RGB}{232,151,218}
\definecolor{nn-softmax-color}{RGB}{251,255,2}
\definecolor{nn-pool-color}{RGB}{238,189,154}
\definecolor{nn-unpool-color}{RGB}{154,189,238}
\definecolor{nn-concat-color}{RGB}{191,132,253}
\definecolor{nn-sum-color}{RGB}{127,255,255}
\definecolor{nn-cu-color}{RGB}{255,216,0}

\tikzstyle{nn-node}=[draw, thick, rounded corners=5pt]
\tikzstyle{nn-teaser}=[minimum height=0.6cm, minimum width=0.6cm]
\tikzstyle{nn-minor}=[minimum height=0.45cm, minimum width=0.45cm, Gray!70, text=Black]
\tikzstyle{nn-edge}=[thick, rounded corners=3pt]
\tikzstyle{nn-arrow}=[->, >=stealth]

\tikzstyle{nn-input}=[fill=nn-input-color]
\tikzstyle{nn-fc}=[fill=nn-fc-color]
\tikzstyle{nn-conv}=[fill=nn-conv-color]
\tikzstyle{nn-bias}=[fill=nn-bias-color]
\tikzstyle{nn-bn}=[fill=nn-bn-color]
\tikzstyle{nn-relu}=[fill=nn-relu-color]
\tikzstyle{nn-softmax}=[fill=nn-softmax-color]
\tikzstyle{nn-pool}=[fill=nn-pool-color]
\tikzstyle{nn-unpool}=[fill=nn-unpool-color]
\tikzstyle{nn-concat}=[fill=nn-concat-color]
\tikzstyle{nn-sum}=[fill=nn-sum-color]
\tikzstyle{nn-cu}=[fill=nn-cu-color]

%% file: color_table2.tex
\begin{tikzpicture}[tight background, scale=0.75, every node/.style={font=\large}]
	\draw[white, fill=csunlabeled, draw=white] (0,0) rectangle (1 * 4, 1) node[pos=0.5] {Void};
	\draw[white, fill=csroad, draw=white] (1 * 4,0) rectangle (2 * 4, 1) node[pos=0.5] {Road};
	\draw[white, fill=cssidewalk, draw=white] (2 * 4,0) rectangle (3 * 4, 1) node[pos=0.5] {Sidewalk};
	\draw[white, fill=csbuilding, draw=white] (3 * 4,0) rectangle (4 * 4, 1) node[pos=0.5] {Building};
	\draw[white, fill=cswall, draw=white] (4 * 4,-0) rectangle (5 * 4, 1) node[pos=0.5] {Wall};
	\draw[black, fill=csfence, draw=white] (5 * 4,-0) rectangle (6 * 4, 1) node[pos=0.5] {Fence};
	\draw[white, fill=cspole, draw=white] (6 * 4,-0) rectangle (7 * 4, 1) node[pos=0.5] {Pole};
	\draw[white, fill=cstrafficlight, draw=white] (7 * 4,-0) rectangle (8 * 4, 1) node[pos=0.5] {Traffic Light};
	\draw[black, fill=cstrafficsign, draw=white] (8 * 4,-0) rectangle (9 * 4, 1) node[pos=0.5] {Traffic Sign};
	\draw[white, fill=csvegetation, draw=white] (9 * 4,-0) rectangle (10 * 4, 1) node[pos=0.5] {Vegetation};

	\draw[black, fill=csterrain, draw=white] (0 * 4,-1) rectangle (1 * 4, 0) node[pos=0.5] {Terrain};
	\draw[white, fill=cssky, draw=white] (1 * 4,-1) rectangle (4 * 2, 0) node[pos=0.5] {Sky};
	\draw[white, fill=csperson, draw=white] (2 * 4,-1) rectangle (3 * 4, 0) node[pos=0.5] {Person};
	\draw[white, fill=csrider, draw=white] (3 * 4,-1) rectangle (4 * 4, 0) node[pos=0.5] {Rider};
	\draw[white, fill=cscar, draw=white] (4 * 4,-1) rectangle (5 * 4, 0) node[pos=0.5] {Car};
	\draw[white, fill=cstruck, draw=white] (5 * 4,-1) rectangle (6 * 4, 0) node[pos=0.5] {Truck};
	\draw[white, fill=csbus, draw=white] (6 * 4,-1) rectangle (7 * 4, 0) node[pos=0.5] {Bus};
	\draw[white, fill=cstrain, draw=white] (7 * 4,-1) rectangle (8 * 4, 0) node[pos=0.5] {Train};
	\draw[white, fill=csmotorcycle, draw=white] (8 * 4,-1) rectangle (9 * 4, 0) node[pos=0.5] {Motorcycle};
	\draw[white, fill=csbicycle, draw=white] (9 * 4,-1) rectangle (10 * 4, 0) node[pos=0.5] {Bicycle};
\end{tikzpicture}

%% file: frrn_paper.bbl
\begin{thebibliography}{10}\itemsep=-1pt

\bibitem{Theano16}
R.~Al-Rfou, G.~Alain, A.~Almahairi, et~al.
\newblock {Theano: A Python framework for fast computation of mathematical
  expressions}.
\newblock abs/1605.02688, 2016.

\bibitem{Badrinarayanan15aARXIV}
V.~Badrinarayanan, A.~Handa, and R.~Cipolla.
\newblock {SegNet: A Deep Convolutional Encoder-Decoder Architecture for Robust
  Semantic Pixel-Wise Labelling}.
\newblock arXiv:1505.07293, 2015.

\bibitem{Badrinarayanan15bARXIV}
V.~Badrinarayanan, A.~Kendall, and R.~Cipolla.
\newblock {SegmentationNet: A Deep Convolutional Encoder-Decoder Architecture
  for Image Segmentation}.
\newblock arXiv:1511.00561, 2015.

\bibitem{Bansal09ICCVW}
M.~Bansal, B.~Matei, H.~Sawhney, S.-H. Jung, and J.~Eledath.
\newblock {Pedestrian Detection with Depth-Guided Structure Labeling}.
\newblock In {\em {ICCV Workshop}}, 2009.

\bibitem{Bergstra13ICML}
J.~Bergstra, D.~Yamins, and D.~D. Cox.
\newblock {Making a Science of Model Search: Hyperparameter Optimization in
  Hundreds of Dimensions for Vision Architectures}.
\newblock In {\em ICML}, 2013.

\bibitem{Chandra16ARXIV}
S.~Chandra and I.~Kokkinos.
\newblock {Fast, Exact and Multi-Scale Inference for Semantic Image
  Segmentation with Deep Gaussian {CRF}s}.
\newblock arXiv:1603.08358, 2016.

\bibitem{HChen16ARXIV}
H.~Chen, Q.~Dou, L.~Yu, and P.~Heng.
\newblock {VoxResNet: Deep Voxelwise Residual Networks for Volumetric Brain
  Segmentation}.
\newblock arXiv:1608.05895, 2016.

\bibitem{Chen15ICLR}
L.~Chen, G.~Papandreou, I.~Kokkinos, K.~Murphy, and A.~L. Yuille.
\newblock {Semantic Image Segmentation with Deep Convolutional Nets and Fully
  Connected {CRF}s}.
\newblock ICLR, 2015.

\bibitem{Chen16ARXIV}
L.~Chen, G.~Papandreou, I.~Kokkinos, K.~Murphy, and A.~L. Yuille.
\newblock {DeepLab: Semantic Image Segmentation with Deep Convolutional Nets,
  Atrous Convolution, and Fully Connected CRFs}.
\newblock arXiv:1606.00915, 2016.

\bibitem{TChen16ARXIV}
T.~Chen, B.~Xu, C.~Zhang, and C.~Guestrin.
\newblock {Training Deep Nets with Sublinear Memory Cost}.
\newblock arXiv:1604.06174, 2016.

\bibitem{Cordts16CVPR}
M.~Cordts, M.~Omran, S.~Ramos, T.~Rehfeld, M.~Enzweiler, R.~Benenson,
  U.~Franke, S.~Roth, and B.~Schiele.
\newblock {The Cityscapes Dataset for Semantic Urban Scene Understanding}.
\newblock In {\em CVPR}, 2016.

\bibitem{Dai15ICCV}
J.~Dai, K.~He, and J.~Sun.
\newblock {BoxSup: Exploiting Bounding Boxes to Supervise Convolutional
  Networks for Semantic Segmentation}.
\newblock In {\em ICCV}, 2015.

\bibitem{Dai2015CVPR}
J.~Dai, K.~He, and J.~Sun.
\newblock {Convolutional Feature Masking for Joint Object and Stuff
  Segmentation}.
\newblock In {\em CVPR}, 2015.

\bibitem{Lasagne15}
S.~Dieleman, J.~Schlüter, C.~Raffel, et~al.
\newblock {Lasagne: First release.}, Aug. 2015.

\bibitem{Ess2009BMVC}
A.~Ess, T.~M{\"u}ller, H.~Grabner, and L.~Van~Gool.
\newblock {Segmentation-Based Urban Traffic Scene Understanding}.
\newblock In {\em BMVC}, 2009.

\bibitem{Farabet13TPAMI}
C.~Farabet, C.~Couprie, L.~Najman, and Y.~LeCun.
\newblock {Learning Hierarchical Features for Scene Labeling}.
\newblock {\em PAMI}, 35(8), 2013.

\bibitem{Floros2012CVPR}
G.~Floros and B.~Leibe.
\newblock Joint 2d-3d temporally consistent semantic segmentation of street
  scenes.
\newblock In {\em CVPR}, pages 2823--2830. IEEE, 2012.

\bibitem{Gaidon16CVPR}
A.~Gaidon, Q.~Wang, Y.~Cabon, and E.~Vig.
\newblock {Virtual Worlds as Proxy for Multi-Object Tracking Analysis}.
\newblock In {\em CVPR}, 2016.

\bibitem{Gastal11TOG}
E.~S.~L. Gastal and M.~M. Oliveira.
\newblock {Domain Transform for Edge-Aware Image and Video Processing}.
\newblock In {\em ACM Trans. Graphics}, 2011.

\bibitem{Ghiasi16ECCV}
G.~Ghiasi and C.~C. Fowlkes.
\newblock {Laplacian Reconstruction and Refinement for Semantic Segmentation}.
\newblock In {\em ECCV}, 2016.

\bibitem{Gould08IJCV}
S.~Gould, J.~Rodgers, D.~Cohen, G.~Elidan, and D.~Koller.
\newblock Multi-class segmentation with relative location prior.
\newblock {\em IJCV}, 80(3):300--316, 2008.

\bibitem{Gruslys16ARXIV}
A.~Gruslys, R.~Munos, I.~Danihelka, M.~Lanctot, and A.~Graves.
\newblock {Memory-Efficient Backpropagation Through Time}.
\newblock arXiv:1606.03401, 2016.

\bibitem{Gu09CVPR}
C.~Gu, J.~J. Lim, P.~Arbel{\'a}ez, and J.~Malik.
\newblock {Recognition using Regions}.
\newblock In {\em CVPR}. IEEE, 2009.

\bibitem{Hariharan14ECCV}
B.~Hariharan, P.~Arbel{\'a}ez, R.~Girshick, and J.~Malik.
\newblock {Simultaneous Detection and Segmentation}.
\newblock In {\em ECCV}. Springer, 2014.

\bibitem{He16CVPR}
K.~He, X.~Zhang, S.~Ren, and J.~Sun.
\newblock {Deep Residual Learning for Image Recognition}.
\newblock In {\em CVPR}, 2016.

\bibitem{He16ECCV}
K.~He, X.~Zhang, S.~Ren, and J.~Sun.
\newblock {Identity Mappings in Deep Residual Networks}.
\newblock In {\em ECCV}, 2016.

\bibitem{Ioffe15ICML}
S.~Ioffe and C.~Szegedy.
\newblock {Batch Normalization: Accelerating Deep Network Training by Reducing
  Internal Covariate Shift}.
\newblock In {\em ICML}, 2015.

\bibitem{Kingma15ICLR}
D.~P. Kingma and J.~Ba.
\newblock {Adam: A Method for Stochastic Optimization}.
\newblock In {\em ICLR}, 2015.

\bibitem{Kohli09IJCV}
P.~Kohli, P.~H. Torr, et~al.
\newblock {Robust Higher Order Potentials for Enforcing Label Consistency}.
\newblock {\em IJCV}, 82(3):302--324, 2009.

\bibitem{Krahenbuhl11NIPS}
P.~Kr{\"{a}}henb{\"{u}}hl and V.~Koltun.
\newblock {Efficient Inference in Fully Connected {CRF}s with {G}aussian Edge
  Potentials}.
\newblock In {\em NIPS}, 2011.

\bibitem{Krizhevsky12NIPS}
A.~Krizhevsky, I.~Sutskever, and G.~Hinton.
\newblock {ImageNet Classification with Deep Convolutional Networks}.
\newblock In {\em NIPS}, 2012.

\bibitem{Kundu14jECCV}
A.~Kundu, Y.~Li, F.~Dellaert, F.~Li, and J.~M. Rehg.
\newblock Joint semantic segmentation and 3d reconstruction from monocular
  video.
\newblock In {\em ECCV}, pages 703--718, 2014.

\bibitem{Ladicky14CVPR}
L.~Ladicky, J.~Shi, and M.~Pollefeys.
\newblock {Pulling things out of perspective}.
\newblock In {\em CVPR}, pages 89--96, 2014.

\bibitem{Lin16CVPR}
G.~Lin, C.~Shen, I.~D. Reid, and A.~van~den Hengel.
\newblock {Efficient piecewise training of deep structured models for semantic
  segmentation}.
\newblock In {\em CVPR}, 2016.

\bibitem{Liu10CVPR}
B.~Liu, S.~Gould, and D.~Koller.
\newblock Single image depth estimation from predicted semantic labels.
\newblock In {\em CVPR}, pages 1253--1260. IEEE, 2010.

\bibitem{Liu2015ICLRW}
W.~Liu, A.~Rabinovich, and A.~C. Berg.
\newblock {ParseNet: Looking Wider to See Better}.
\newblock 2015.

\bibitem{Liu15ICCV}
Z.~Liu, X.~Li, P.~Luo, C.~C. Loy, and X.~Tang.
\newblock {Semantic Image Segmentation via Deep Parsing Network}.
\newblock In {\em ICCV}, 2015.

\bibitem{Long15CVPR}
J.~Long, E.~Shelhamer, and T.~Darrell.
\newblock {Fully Convolutional Networks for Semantic Segmentation}.
\newblock In {\em CVPR}, 2015.

\bibitem{Mostajabi15CVPR}
M.~Mostajabi, P.~Yadollahpour, and G.~Shakhnarovich.
\newblock {Feedforward Semantic Segmentation With Zoom-Out Features}.
\newblock In {\em CVPR}, 2015.

\bibitem{Newell16ECCV}
A.~Newell, K.~Yang, and J.~Deng.
\newblock {Stacked Hourglass Networks for Human Pose Estimation}.
\newblock In {\em ECCV}, 2016.

\bibitem{Noh15ICCV}
H.~Noh, S.~Hong, and B.~Han.
\newblock {Learning Deconvolution Network for Semantic Segmentation}.
\newblock In {\em ICCV}, 2015.

\bibitem{Osep16ICRA}
A.~O\v{s}ep, A.~Hermans, F.~Engelmann, D.~Klostermann, M.~Mathias, and
  B.~Leibe.
\newblock {Multi-Scale Object Candidates for Generic Object Tracking in Street
  Scenes}.
\newblock In {\em ICRA}, 2016.

\bibitem{Papandreou15ICCV}
G.~Papandreou, L.~Chen, K.~P. Murphy, and A.~L. Yuille.
\newblock {Weakly-and Semi-Supervised Learning of a Deep Convolutional Network
  for Semantic Image Segmentation}.
\newblock In {\em ICCV}, 2015.

\bibitem{Paszke16ARXIV}
A.~Paszke, A.~Chaurasia, S.~Kim, and E.~Culurciello.
\newblock {ENet: A Deep Neural Network Architecture for Real-Time Semantic
  Segmentation}.
\newblock arXiv:1606.02147, 2016.

\bibitem{Ros16CVPR}
G.~Ros, L.~Sellart, J.~Materzynska, D.~Vazquez, and A.~Lopez.
\newblock {The SYNTHIA Dataset: A Large Collection of Synthetic Images for
  Semantic Segmentation of Urban Scenes}.
\newblock In {\em CVPR}, 2016.

\bibitem{Rumelhart86Nature}
D.~E. Rumelhart, G.~E. Hinton, and R.~J. Williams.
\newblock {Learning representations by back-propagating errors}.
\newblock {\em Nature}, 323, 1986.

\bibitem{Russakovsky15IJSCV}
O.~Russakovsky, J.~Deng, H.~Su, J.~Krause, S.~Satheesh, S.~Ma, Z.~Huang,
  A.~Karpathy, A.~Khosla, M.~S. Bernstein, A.~C. Berg, and F.~Li.
\newblock {ImageNet Large Scale Visual Recognition Challenge}.
\newblock {\em IJCV}, 115(3), 2015.

\bibitem{Schwing15ARXIV}
A.~G. Schwing and R.~Urtasun.
\newblock {Fully Connected Deep Structured Networks}.
\newblock arXiv:1503.02351, 2015.

\bibitem{Shotton08CVPR}
J.~Shotton, M.~Johnson, and R.~Cipolla.
\newblock {Semantic texton forests for Image categorization and segmentation}.
\newblock In {\em CVPR}, 2008.

\bibitem{Simonyan15ICLR}
K.~Simonyan and A.~Zisserman.
\newblock {Very Deep Convolutional Networks for Large-Scale Image Recognition}.
\newblock In {\em ICLR}, 2015.

\bibitem{Szegedy15CVPR}
C.~Szegedy, W.~Liu, Y.~Jia, P.~Sermanet, S.~Reed, D.~Anguelov, D.~Erhan,
  V.~Vanhoucke, and A.~Rabinovich.
\newblock {Going Deeper with Convolutions}.
\newblock In {\em CVPR}, 2015.

\bibitem{Wu16ARXIV}
Z.~Wu, C.~Shen, and A.~v.~d. Hengel.
\newblock {Bridging Category-level and Instance-level Semantic Image
  Segmentation}.
\newblock arXiv:1605.06885, 2016.

\bibitem{xiao09ICCV}
J.~Xiao and L.~Quan.
\newblock Multiple view semantic segmentation for street view images.
\newblock In {\em ICCV}. IEEE, 2009.

\bibitem{yan15CVPR}
J.~Yan, Y.~Yu, X.~Zhu, Z.~Lei, and S.~Z. Li.
\newblock {Object Detection by Labeling Superpixels}.
\newblock In {\em CVPR}, 2015.

\bibitem{Yu16ICLR}
F.~Yu and V.~Koltun.
\newblock {Multi-Scale Context Aggregation by Dilated Convolutions}.
\newblock In {\em ICLR}, 2016.

\bibitem{Zeiler11CVPR}
M.~D. Zeiler, G.~W. Taylor, and R.~Fergus.
\newblock {Adaptive Deconvolutional Networks for Mid and High Level Feature
  Learning}.
\newblock In {\em CVPR}, 2011.

\bibitem{Zheng15ICCV}
S.~Zheng, S.~Jayasumana, B.~Romera{-}Paredes, V.~Vineet, Z.~Su, D.~Du,
  C.~Huang, and P.~H.~S. Torr.
\newblock {Conditional Random Fields as Recurrent Neural Networks}.
\newblock In {\em ICCV}, 2015.

\bibitem{Zhu15CVPR}
Y.~Zhu, R.~Urtasun, R.~Salakhutdinov, and S.~Fidler.
\newblock {segDeepM: Exploiting Segmentation and Context in Deep Neural
  Networks for Object Detection}.
\newblock In {\em CVPR}, 2015.

\end{thebibliography}


\begin{thebibliography}{1}\itemsep=-1pt

\bibitem{Ghiasi16ECCV}
G.~Ghiasi and C.~C. Fowlkes.
\newblock {Laplacian Reconstruction and Refinement for Semantic Segmentation}.
\newblock In {\em ECCV}, 2016.

\bibitem{Yu16ICLR}
F.~Yu and V.~Koltun.
\newblock {Multi-Scale Context Aggregation by Dilated Convolutions}.
\newblock In {\em ICLR}, 2016.

\end{thebibliography}
